\documentclass[runningheads]{llncs}

\usepackage{eccv}

\usepackage{eccvabbrv}

\usepackage{graphicx}
\usepackage{booktabs}
\usepackage{multirow}
\usepackage{colortbl}
\usepackage{xcolor}
\usepackage{wrapfig}

\usepackage[accsupp]{axessibility}  

\usepackage{hyperref}

\usepackage{orcidlink}

\begin{document}

\title{HyLaR: Hybrid Latent Reasoning with Decoupled Policy Optimization} 

\titlerunning{Hybrid Latent Reasoning with Decoupled Policy Optimization}

\author{Tao Cheng\inst{1}\orcidlink{0009-0006-7905-9807} \and Shi-Zhe Chen\inst{1}\textsuperscript{,*,\textdagger}\orcidlink{0009-0002-4677-4581} \and Hao Zhang\inst{1}\orcidlink{0009-0004-7469-6896} \and Yixin Qin\inst{1}\orcidlink{0009-0004-1067-062X} \\ Jinwen Luo\inst{2}\orcidlink{0009-0003-8663-3836} \and Zheng Wei\inst{1}\textsuperscript{,*}\orcidlink{0009-0008-1684-436X}}

\authorrunning{T.~Cheng et al.}

\institute{$^1$ Tencent PCG, China \quad $^2$ Tencent CSIG, China \\ \email{\{elvistcheng,shizhechen,jeffhzhang,wadeqin,jamsluo,hemingwei\}@tencent.com}}

\maketitle
\renewcommand{\thefootnote}{\fnsymbol{footnote}}
\footnotetext[1]{Corresponding author.}
\footnotetext[4]{Project lead.}
\renewcommand{\thefootnote}{\arabic{footnote}}

\begin{abstract}
  Chain-of-Thought (CoT) reasoning significantly elevates the complex problem-solving capabilities of multimodal large language models (MLLMs). However, adapting CoT to vision typically discretizes signals to fit LLM inputs, causing early semantic collapse and discarding fine-grained details. While external tools can mitigate this, they introduce a rigid bottleneck, confining reasoning to predefined operations. Although recent latent reasoning paradigms internalize visual states to overcome these limitations, optimizing the resulting hybrid discrete-continuous action space remains challenging. In this work, we propose HyLaR (Hybrid Latent Reasoning), a framework that seamlessly interleaves discrete text generation with continuous visual latent representations. Specifically, following an initial cold-start supervised fine-tuning (SFT), we introduce DePO (Decoupled Policy Optimization) to enable effective reinforcement learning within this hybrid space. DePO decomposes the policy gradient objective, applying independent trust-region constraints to the textual and latent components, alongside an exact closed-form von Mises-Fisher (vMF) KL regularizer. Extensive experiments demonstrate that HyLaR outperforms standard MLLMs and state-of-the-art latent reasoning approaches across fine-grained perception and general multimodal understanding benchmarks. Code is available at \url{https://github.com/EthenCheng/HyLaR}.

  \keywords{Multimodal Large Language Models \and Visual Latent Reasoning \and Policy Optimization}
\end{abstract}

\section{Introduction}
\label{sec:intro}

\begin{figure}[htbp]
  \centering
  \includegraphics[width=0.85\textwidth]{images/inference.pdf}  
  \caption{HyLaR vs.\ two reasoning paradigms. \textbf{(A) Text-only CoT:} visual grounding errors and redundant steps. \textbf{(B) Think-with-Images:} external tools cause unstable invocations and latency. \textbf{(C) HyLaR (ours):} refines latent think tokens in latent space, preserving fine-grained visual evidence.}
  \label{fig:intro_cmp}                                
\end{figure}

The integration of explicit Chain-of-Thought (CoT) reasoning has fundamentally transformed how multimodal large language models (MLLMs) approach intricate vision-language tasks~\cite{qwen3-VL, internvl3, glmv, gemini-2.5, vision-r1, mmcot}. However, most prevailing MLLMs suffer from a critical architectural bottleneck: \textit{early semantic collapse}. As illustrated in Fig.~\ref{fig:intro_cmp}(A), standard text-only CoT forces high-bandwidth, continuous visual signals to be prematurely compressed into discrete text tokens. This early discretization inevitably discards fine-grained visual evidence that is difficult to verbalize, leaving the model to rely on linguistic priors rather than grounded visual facts.

To mitigate this, recent studies have explored two primary alternatives. The first is the ``Think-with-Images'' paradigm (Fig.~\ref{fig:intro_cmp}(B)), which relies on external tools to re-perceive the image, yet introduces rigid bottlenecks and inference latency~\cite{deepeyes, deepeyesv2, thyme}. The second alternative, including recent pioneering works, such as LVR~\cite{lvr-latent}, SkiLa~\cite{sketch} and Monet~\cite{monet}, shifts reasoning into a continuous latent space to preserve visual fidelity. While promising, optimizing the resulting hybrid discrete-continuous action space remains profoundly challenging. Current approaches predominantly relying on supervised fine-tuning (SFT) or vanilla reinforcement learning (RL) often yield sub-optimal optimization, as conventional Gaussian assumptions and uniform clipping fundamentally mismatch the native hyperspherical geometry of MLLMs and the distinct variance of hybrid actions.

To overcome these challenges, we propose \textbf{HyLaR} (\textbf{Hy}brid \textbf{La}tent \textbf{R}easoning), an elegant and highly effective framework that seamlessly interleaves discrete text generation with continuous visual latent representations (Fig.~\ref{fig:intro_cmp}(C)). Following a straightforward cold-start SFT phase that teaches the model to dynamically alternate between logical text tokens and continuous visual working memory, we focus on unlocking the full potential of this hybrid space through reinforcement learning.
The core innovation of this work is \textbf{DePO} (\textbf{De}coupled \textbf{P}olicy \textbf{O}ptimization), an RL algorithm tailored specifically for hybrid action spaces. Standard policy optimization applies uniform constraints and assumes a flat continuous space, which fails in our setting due to two critical mismatches. First, the importance sampling ratios of discrete tokens and continuous vectors exhibit vastly different variance characteristics. DePO resolves this via \textit{decoupled trust-region clipping}, applying position-specific clipping ranges to stabilize continuous updates without hindering text convergence. 
Second, and more importantly, rather than relying on standard Gaussian distributions that imply a Euclidean geometry, we explicitly model the continuous latent policy using the vMF distribution~\cite{wang2020understanding,davidson2018hyperspherical,gauthier2024exploring}. Because layer-normalized LLM representations inherently reside on a high-dimensional hypersphere, the vMF formulation perfectly aligns with this native geometry. This elegant spherical modeling gracefully translates intractable probability estimations and KL divergences into exact, closed-form cosine distances. Consequently, DePO renders the latent RL pipeline remarkably simpler, more geometrically rigorous, and easier to implement than prior arts.

Our main contributions are summarized as follows:
\begin{itemize}
    \item We present HyLaR, a hybrid reasoning framework that overcomes early semantic collapse by interleaving discrete textual logic with continuous visual latent representations.
    \item We introduce DePO, a novel reinforcement learning algorithm for hybrid action spaces. By identifying the geometric and variance mismatches in standard RL, DePO leverages hyperspherical vMF modeling and decoupled clipping to achieve highly effective and exact latent policy optimization.
    \item Extensive experiments validate HyLaR's superiority in high-resolution perception and complex visual reasoning. Furthermore, rigorous ablations on latent scaling, geometric modeling, and decoupled clipping firmly establish the stability and necessity of our framework.
\end{itemize}

\section{Related Work}
\label{relatedwork}

\subsection{Explicit Reasoning in MLLMs}
A substantial body of work has explored visual reasoning in vision-language models. Early approaches~\cite{llava-onevision, Vl-rethinker, Visual_programming_CVPR, mmcot} primarily relied on textual Chain-of-Thought (CoT) prompting, where the model performed a single inference after encoding both the image and the question. However, subsequent studies revealed a fundamental disconnect: the reasoning process often fails to properly leverage perceptual outputs, leading to scenarios where models correctly perceive visual information yet still produce erroneous answers. To address these limitations, recent work has shifted toward ``Think-with-Images'' paradigms~\cite{deepeyes, deepeyesv2, thyme}, which employ explicit tool invocations to zoom in on or crop specific regions of interest. Other works incorporate generative visual modules~\cite{showo, deem, think_diffuse} to produce pixel-level intermediate outputs. While many of these approaches have successfully employed reinforcement learning (RL) to optimize discrete textual trajectories or tool invocations~\cite{deepeyes}, they still fundamentally rely on externalized visual processing, introducing tool invocation errors, inference latency, and rigid operational bottlenecks.

\subsection{Latent Space Reasoning}
To avoid early discretization and external overhead, recent studies have increasingly shifted toward latent space reasoning by replacing discrete tokens with continuous embedding representations during autoregressive generation~\cite{soft-thinking, codi, COCONUT, simcot, yu2026latent}. This paradigm exploits the richer information encoded in latent vectors to enhance reasoning flexibility and compress lengthy reasoning chains. 

Beyond language-only models, latent reasoning has been actively extended to MLLMs through various strategies. For instance, Ray et al.~\cite{mull-tokens} introduce modality-agnostic latent tokens as an internal scratchpad to facilitate free-form reasoning. Other approaches, such as LVR~\cite{lvr-latent} and Mirage~\cite{machine}, propose aligning generated latent embeddings with those of auxiliary images. While effective, these designs involve inherent trade-offs: Mirage~\cite{machine} further compresses image embeddings via mean pooling before alignment, whereas LVR~\cite{lvr-latent} focuses primarily on cropped image regions, limiting its capacity to encode visual operations over the entire image. To preserve global context, Laser~\cite{Laser} enforces a ``forest-before-trees'' cognitive hierarchy via dynamic windowed alignment, maintaining a probabilistic superposition of global features to prevent premature semantic collapse. From a training perspective, SkiLa~\cite{sketch} offers a straightforward SFT approach similar to LVR, yet it lacks RL optimization. Conversely, Monet~\cite{monet} presents a highly comprehensive framework combining a complex three-stage SFT pipeline with subsequent RL, though its intricate design can be challenging to implement and susceptible to accumulated training bias.

While RL can further unlock hybrid latent reasoning, standard methods are not fully aligned with the hyperspherical geometry and hybrid action variance of MLLM representations, motivating our Decoupled Policy Optimization (DePO).

\begin{figure}[htbp]
  \centering
  \includegraphics[width=0.85\textwidth]{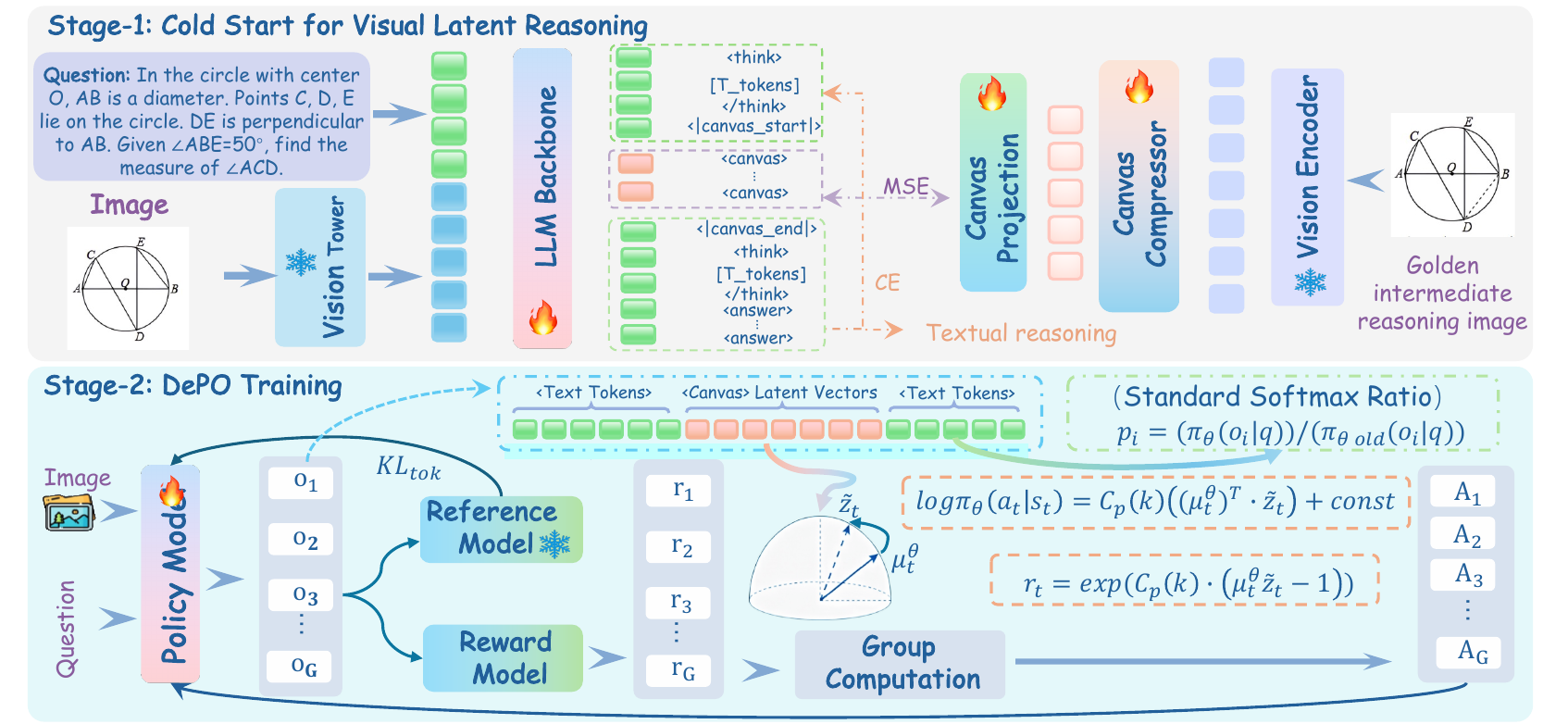}  
  \caption{Overview of the \textbf{HyLaR} framework. \textbf{Stage-I (SFT):} jointly optimizes text via cross-entropy ($\mathcal{L}_{CE}$) and aligns hidden states with compressed ground-truth canvases via MSE ($\mathcal{L}_{Canvas}$). \textbf{Stage-II (DePO):} refines the hybrid trajectory with RL---text via probability ratios, latent vectors via vMF cosine similarity on a hypersphere.}
  \label{fig:trainpipline}                                
\end{figure}

\section{Method}
\label{method}

In this section, we present \textbf{HyLaR} (\textbf{Hy}brid \textbf{La}tent \textbf{R}easoning), a novel framework that empowers MLLMs to seamlessly interleave discrete textual reasoning with continuous visual latent representations. We first provide an overview of the HyLaR architecture, including its hybrid generation mechanism and the distinct paradigms for training and inference (\S\ref{sec:overview}). Subsequently, we detail the two-stage training methodology: the cold-start SFT with canvas compression and alignment strategy (\S\ref{sec:sft}), followed by the \textbf{DePO} (\textbf{De}coupled \textbf{P}olicy \textbf{O}ptimization) to further unlock and stabilize latent reasoning capabilities via RL (\S\ref{sec:depo}).

\subsection{Overview of the HyLaR Framework}
\label{sec:overview}
HyLaR is built upon a standard MLLM architecture, but fundamentally extends the traditional discrete decoding space into a hybrid discrete-continuous action space. To achieve this, we introduce specialized control tokens, \texttt{<|canvas\_start|>} and \texttt{<|canvas\_end|>}, to explicitly bound the visual reasoning process. During inference, when the model requires fine-grained spatial or visual deduction, it transitions into the canvas mode. In this mode, textual generation is suspended, and the hidden state from the previous layer is recurrently fed back as the input embedding for the next step, bypassing the discrete token vocabulary. This continuous latent recursion acts as an internal visual working memory and continues autonomously until the \texttt{<|canvas\_end|>} token is emitted or reaches the maximum canvas length budget, effectively enabling deep visual thinking while avoiding the substantial latency of pixel-level image generation.

HyLaR is trained in two stages. We first design a cold-start SFT phase that teaches the model to alternate between text tokens and latent canvas steps, supervising the latter with compressed ground-truth canvases instead of high-resolution images for efficiency (detailed in \S\ref{sec:sft}); crucially, this auxiliary Canvas Compressor is entirely discarded after SFT, ensuring zero architectural overhead at inference. Building on this alignment, we further refine the hybrid generation via Decoupled Policy Optimization (DePO), which applies independent trust-region constraints to optimize reasoning trajectories against task-specific rewards without any explicit intermediate visual supervision (\S\ref{sec:depo}).

\subsection{Stage I: Supervised Fine-Tuning}
\label{sec:sft}
The supervised fine-tuning (SFT) stage equips the MLLM to seamlessly alternate between textual reasoning and latent visual thinking by approximating the semantic representations of ground-truth intermediate canvases.

\paragraph{\textup{\textbf{Canvas Compression and Injection.}}} During training, ground-truth canvas images are first processed by a frozen SigLIP2 encoder~\cite{tschannen2025siglip} to extract $P=729$ patch tokens. To avoid the exorbitant sequence length caused by these raw high-resolution patches, a learnable cross-attention compressor is adopted. Configured with $L=2$ layers and $N=16$ query tokens, the compressor attentively aggregates these patches into $N$ compact canvas embeddings $\mathbf{e}$. Within the training reasoning trajectory, original intermediate images are replaced by \texttt{<canvas>} placeholders bounded by \texttt{<|canvas\_start|>} and \texttt{<|canvas\_end|>}, into which the compressed target embeddings $\mathbf{e}$ are directly injected via masked scattering.

\paragraph{\textup{\textbf{Joint End-to-End Optimization.}}} The model is jointly trained to predict both the next discrete text tokens and the next continuous latent canvas embeddings. We apply the standard autoregressive cross-entropy loss ($\mathcal{L}_{CE}$) for text generation. Simultaneously, we optimize an MSE-based canvas prediction loss:
\begin{equation}
    \mathcal{L}_{\text{Canvas}} = \frac{1}{|\mathcal{S}|} \sum_{t \in \mathcal{S}} \left\| \mathbf{h}_t - \mathbf{e}_{t+1} \right\|_2^2,
    \label{eq:canvas_loss}
\end{equation}
where $\mathbf{h}_t$ is the LLM's hidden state at the canvas position $t \in \mathcal{S}$. Crucially, we do not detach the target embeddings $\mathbf{e}_{t+1}$ during backpropagation. This formulation allows gradients to flow end-to-end from $\mathcal{L}_{Canvas}$ into both the LLM backbone and the external Canvas Compressor, actively pulling the extracted visual features into the LLM's native semantic space. The overall SFT objective is defined as $\mathcal{L}_{\text{SFT}} = \mathcal{L}_{\text{CE}} + \lambda \mathcal{L}_{\text{Canvas}}$.

\subsection{Stage II: Reinforcement Learning with DePO}
\label{sec:depo}

\paragraph{\textup{\textbf{Limitations of Standard RL Objectives.}}}
Standard reinforcement learning objectives, such as those used in PPO~\cite{schulman2017proximal}, GRPO~\cite{grpo}, and DAPO~\cite{yu2025dapo}, typically apply a shared surrogate loss with a uniform clipping rule and sample-based KL regularization across all action steps. While this design is effective for purely discrete text generation, it is not well calibrated for our hybrid reasoning trajectory, where discrete tokens and continuous latent states exhibit fundamentally different statistical and geometric properties. In particular, directly treating these two action types in the same way leads to two key limitations.

\begin{wrapfigure}[20]{r}{0.4\textwidth}
    \centering
    \includegraphics[width=0.4\textwidth]{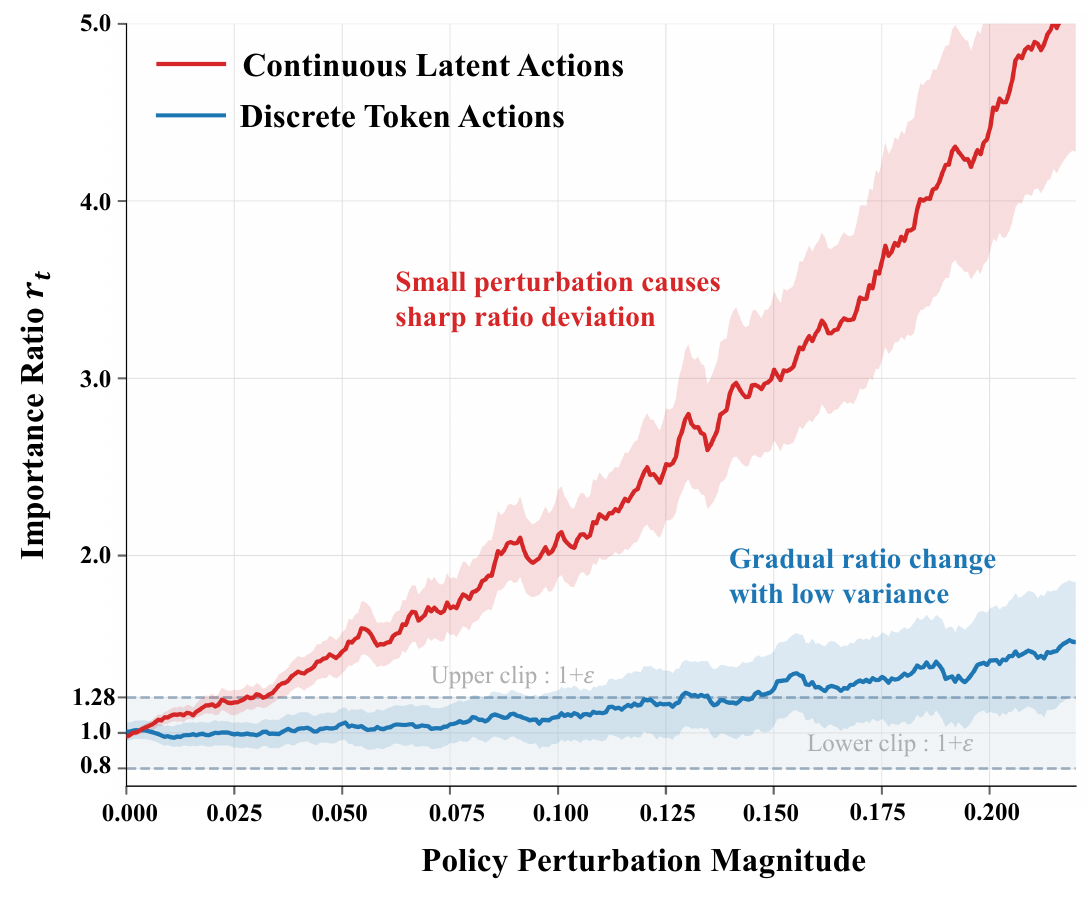}
    \caption{Importance ratio $r_t$ vs.\ policy perturbation magnitude $\lVert\theta - \theta_{\mathrm{old}}\rVert_2 / \lVert\theta_{\mathrm{old}}\rVert_2$ for discrete token vs.\ continuous latent actions. Latent ratios deviate far more sharply under small policy updates, motivating tighter clipping for latent positions.}
    \label{fig:ratio_mismatch}
\end{wrapfigure}

\textbf{(1)~Optimization mismatch between discrete and continuous actions.} The importance ratio $r_t = \pi_\theta(a_t \mid s_t) / \pi_{\theta_{\mathrm{old}}}(a_t \mid s_t)$ behaves differently in discrete token spaces and in high-dimensional continuous latent spaces. For discrete text generation, action probabilities are defined over a normalized categorical distribution, and token-level updates are typically relatively well behaved under standard clipping schemes. By contrast, for continuous latent actions, the log-density often depends on distances or directional deviations in a high-dimensional space, so even small policy changes can induce disproportionately large changes in likelihood ratios (see Fig.~\ref{fig:ratio_mismatch}). This issue is further amplified by autoregressive latent recursion: each latent action is fed into subsequent reasoning steps, so small perturbations can alter future states and accumulate along the rollout. As a result, a single clipping range or regularization strength that is adequate for token updates may be too loose or poorly calibrated for latent updates, leading to unstable optimization.

\textbf{(2)~Geometric mismatch in continuous latent policy modeling.} Standard continuous RL typically assumes a Gaussian action space, which implies a flat Euclidean geometry. However, modern MLLMs heavily employ normalization layers (e.g., RMSNorm~\cite{rmsnorm}) that inherently constrain hidden states to a high-dimensional hyperspherical manifold~\cite{wang2020understanding}. As highlighted by prior representation learning studies~\cite{davidson2018hyperspherical}, applying sample-based Euclidean Gaussian estimation to these hyperspheres suffers from severe geometric distortion. In high-dimensional spaces, this mismatch results in prohibitively high sampling variance for the KL penalty, failing to effectively regularize the policy.

To address these issues, we propose \emph{Decoupled Policy Optimization} (DePO), which explicitly models continuous latent reasoning steps with a hyperspherical policy, while decoupling stable rollout execution from stochastic policy optimization.

\paragraph{\textup{\textbf{Hyperspherical vMF Modeling.}}}
We model continuous latent actions using the von Mises--Fisher (vMF) distribution, which is the natural probability distribution on the unit hypersphere and has recently shown promise in continuous reinforcement learning. In our setting, the model performs latent recursion between the \texttt{<|canvas\_start|>} and \texttt{<|canvas\_end|>} delimiters:
\begin{equation}
\mathbf{h}_t^\theta = f_\theta(\mathbf{h}_{t-1}^\theta),
\end{equation}
where $\mathbf{h}_t^\theta \in \mathbb{R}^D$ denotes the hidden state at step $t$. This induces a hybrid action space in which, at step $t$, the action $a_t$ is either a discrete token $a_t \in \mathcal{V}$ or a continuous latent vector $a_t \in \mathbb{S}^{D-1}$.
For discrete token positions, the policy follows a standard categorical distribution over the vocabulary $\mathcal{V}$:
\begin{equation}
\pi_\theta(a_t \mid s_t) = \mathrm{softmax}(\mathbf{W}_m \mathbf{h}_t^\theta)_{a_t},
\end{equation}
where $\mathbf{W}_m$ denotes the output projection matrix.
For continuous latent positions, a categorical policy is no longer applicable because the action space is continuous. Instead, we define the latent policy density of the vMF policy on the hypersphere as
\begin{equation}
\pi_\theta(a_t \mid s_t) = C_D(\kappa)\exp\left(\kappa (\boldsymbol{\mu}_t^\theta)^\top \mathbf{\tilde z}_t\right),
\qquad \mathbf{\tilde z}_t \in \mathbb{S}^{D-1},
\end{equation}
where $\boldsymbol{\mu}_t^\theta \in \mathbb{S}^{D-1}$ is the $\ell_2$-normalized hidden state $\mathbf{h}_t^\theta$, $\kappa > 0$ is a fixed concentration parameter and $C_D(\kappa)$ is the vMF normalization constant in dimension $D$.

Combining the discrete and continuous cases, we write the unified policy log-probability as
\begin{equation}
\log \pi_\theta(a_t \mid s_t) =
\begin{cases}
\log \pi_\theta(a_t \mid s_t), & a_t \in \mathcal{V}, \\[6pt]
\log C_D(\kappa) + \kappa (\boldsymbol{\mu}_t^\theta)^\top \mathbf{\tilde z}_t, & a_t \in \mathbb{S}^{D-1}.
\end{cases}
\label{eq:unified_logprob}
\end{equation}
Since $\kappa$ is fixed, the normalization term $\log C_D(\kappa)$ cancels when computing policy ratios between the old and new policies.

\paragraph{\textup{\textbf{Decoupled Surrogate Objectives.}}}
Because the unified log-probability in Eq.~\eqref{eq:unified_logprob} places
discrete token actions and continuous latent actions on the same sequence, the
importance-sampling ratio $r_t$ can exhibit very different magnitudes at
text positions versus latent positions. In particular, small directional shifts
of the hidden-state $\mu_t^\theta$ can cause disproportionately
large swings in the vMF ratio at latent positions, destabilizing training.

To address this, we partition each response into the set of \emph{text
positions} $\mathcal{Z}$ and \emph{latent positions} $\mathcal{S}$, and apply
the dual-clip PPO surrogate independently to each set with separate clipping
ranges:
\begin{equation}
    \mathcal{L}_{\text{PPO}}(\theta) = \mathcal{L}_{\text{tok}}(\theta \mid \mathcal{Z}; \epsilon_l^{\text{tok}}, \epsilon_h^{\text{tok}}) + \alpha \mathcal{L}_{\text{lat}}(\theta \mid \mathcal{S}; \epsilon_l^{\text{lat}}, \epsilon_h^{\text{lat}}),
    \label{eq:decoupled_ppo}
\end{equation}
where $\theta$ denotes the model parameters, and $\mathcal{L}_{\text{tok}}(\theta \mid \mathcal{Z}; \epsilon_l^{\text{tok}}, \epsilon_h^{\text{tok}})$ represents the dual-clip PPO objective evaluated over the token positions $\mathcal{Z}$. Here, $\epsilon_l^{\text{tok}}$ and $\epsilon_h^{\text{tok}}$ are the asymmetric lower and upper clip thresholds, respectively. The latent loss $\mathcal{L}_{\text{lat}}$ is defined analogously over latent positions $\mathcal{S}$, with $\alpha$ balancing the two terms. Crucially, we use a substantially tighter clipping range for latent positions ($\epsilon^{\text{lat}} \ll \epsilon^{\text{tok}}$) to counteract the higher ratio volatility inherent to continuous vMF actions. We set $\epsilon_l^{\text{tok}}=0.2$, $\epsilon_h^{\text{tok}}=0.28$, and $\epsilon_l^{\text{lat}}=\epsilon_h^{\text{lat}}=0.05$ in our experiments.

\paragraph{\textup{\textbf{Closed-form vMF KL Regularization.}}}
To prevent policy degradation, we apply position-specific KL penalties. For text token positions, we utilize the standard sample-based KL penalty against the reference policy $\pi_\text{ref}$.
For latent positions, the vMF assumption allows us to completely bypass the high-variance sample-based estimation. As proven in supplemental materials, the exact KL divergence between two vMF distributions sharing the same concentration $\kappa$ elegantly reduces to a scaled cosine distance. Thus, the latent KL loss is computed in closed-form:
\begin{equation}
    \mathcal{L}_\text{KL}^\text{lat} = \frac{1}{|\mathcal{S}|} 
    \sum_{t \in \mathcal{S}} \mathbf{W}_\kappa \cdot 
    \left( 1 - \cos\left(\boldsymbol{\mu}_t^\theta,\, \mathbf{\tilde z}_t\right)\right),
    \label{eq:latent_kl_loss}
\end{equation}
where $\mathbf{\mu}_t^\theta$ is the current policy's normalized hidden state (retaining gradients), $\mathbf{\tilde z}_t$ is the previously sampled latent vector from the old policy, and $\mathbf{W}_\kappa$ is a constant weight derived from $\kappa$. This objective provides an exact, zero-variance KL estimate that natively respects the spherical geometry of LLM representations.

\paragraph{\textup{\textbf{Total Objective.}}}
The overall training objective combines the decoupled surrogate losses with the position-specific KL penalties:
\begin{equation}
    \mathcal{L}_\text{total} = \mathcal{L}_\text{PPO} + \beta_\text{tok} \mathcal{L}_\text{KL}^\text{tok} + \beta_\text{lat} \mathcal{L}_\text{KL}^\text{lat},
    \label{eq:total_loss}
\end{equation}
where $\beta_\text{tok}$ and $\beta_\text{lat}$ control the regularization strength for text tokens and latent tokens, respectively. In practice, we set $\beta_\text{tok} = 0.01$ and $\beta_\text{lat} = 0.005$.

\section{Experiments}
\label{sec:experiments}
We conduct extensive experiments across multiple benchmarks to evaluate our method. Specifically, our experiments aim to: (1) demonstrate the effectiveness of the proposed approach against competitive baselines, (2) analyze the contribution of individual components through ablation studies, and (3) provide insights into visual latent reasoning.

\begin{table*}[t]
\centering
\caption{Results on High-Resolution Image Perception and Visual Search Benchmarks. The best-performing latent reasoning model is highlighted in \textbf{bold}. ($^*$Reproduced via our evaluation pipeline for fair comparison; original reported scores are in \textcolor{lightgray}{gray}.)}
\label{tab:main_benchmark}
\resizebox{\textwidth}{!}{%
\begin{tabular}{l ccc ccc ccc}
\toprule
\multirow{2}{*}{\textbf{Methods}} & \multicolumn{3}{c}{\textbf{V*}} & \multicolumn{3}{c}{\textbf{HRBench-4K}} & \multicolumn{3}{c}{\textbf{HRBench-8K}} \\
\cmidrule(lr){2-4} \cmidrule(lr){5-7} \cmidrule(lr){8-10}
& Overall & Attribute & Spatial & Overall & FSP & FCP & Overall & FSP & FCP \\
\midrule
\rowcolor{orange!10} \multicolumn{10}{c}{\textit{Proprietary Model}} \\
GPT-4o~\cite{gpt4o} & 67.5 & 72.2 & 60.5 & 59.0 & 70.0 & 48.0 & 55.5 & 62.0 & 49.0 \\
Gemini-3-Flash~\cite{gemini3flash2025} & 86.4 & - & - & 87.9 & - & - & 85.0 & - & - \\
\midrule
\rowcolor{blue!8} \multicolumn{10}{c}{\textit{Open-Source Model}} \\
LLaVA-OneVision-7B~\cite{llava-onevision} & 71.25 & 73.48 & 67.89 & 62.50 & 74.25 & 50.75 & 58.25 & 67.50 & 49.00 \\
Qwen2.5-VL-7B~\cite{Qwen2.5-VL} & 76.44 & 77.39 & 75.00 & 68.00 & 80.25 & 55.75 & 63.75 & 73.75 & 53.75 \\
InternVL3-8B~\cite{internvl3} & 70.2 & 67.8 & 73.7 & 70.0 & 78.8 & 61.3 & 69.3 & 78.8 & 59.8 \\
\midrule
\rowcolor{purple!10} \multicolumn{10}{c}{\textit{Thinking-with-Images Agent Model}} \\
ZoomEye~\cite{zoomeye} & 79.85 & 80.52 & 78.82 & 68.75 & 81.25 & 56.25 & 64.50 & 75.00 & 54.00 \\
Thyme~\cite{thyme} & 82.2 & 83.5 & 80.3 & 77.0 & 91.0 & 63.0 & 72.0 & 86.5 & 57.5 \\
DeepEyes~\cite{deepeyes} & 85.6 & - & - & 75.1 & - & - & 72.6 & - & - \\
DeepEyesV2~\cite{deepeyesv2} & 81.8 & - & - & 77.9 & - & - & 73.8 & - & - \\
\midrule
\rowcolor{green!10} \multicolumn{10}{c}{\textit{Visual-Latent Model}} \\
LVR~\cite{lvr-latent} & 80.6 & 81.7 & 79.0 & - & - & - & - & - & - \\
Laser~\cite{Laser} & - & - & - & 72.50 & - & - & - & - & - \\
SkiLa$^*$~\cite{sketch} & 78.53 & - & - & 72.12 & - & - & 66.5 & - & - \\
\textcolor{lightgray}{SkiLa~\cite{sketch}} & \textcolor{lightgray}{84.3} & \textcolor{lightgray}{-} & \textcolor{lightgray}{-} & \textcolor{lightgray}{72.0} & \textcolor{lightgray}{-} & \textcolor{lightgray}{-} & \textcolor{lightgray}{66.5} & \textcolor{lightgray}{-} & \textcolor{lightgray}{-} \\
Monet$^*$~\cite{monet} & 80.1 & 81.73 & 77.63 & 67.37 & 78.25 & \textbf{56.50} & 64.37 & 74.25 & \textbf{54.49} \\
\textcolor{lightgray}{Monet~\cite{monet}} & \textcolor{lightgray}{83.25} & \textcolor{lightgray}{83.48} & \textcolor{lightgray}{82.89} & \textcolor{lightgray}{71.00} & \textcolor{lightgray}{85.25} & \textcolor{lightgray}{56.75} & \textcolor{lightgray}{68.00} & \textcolor{lightgray}{79.75} & \textcolor{lightgray}{56.25} \\
\textbf{HyLaR-SFT (Ours)} & 80.63 & 81.73 & 78.95 & 71.50 & 93.00 & 50.00 & 67.19 & 87.01 & 47.11 \\
\textbf{HyLaR-7B (Ours)} & \textbf{83.77} & \textbf{82.61} & \textbf{85.53} & \textbf{75.00} & \textbf{93.75} & 56.25 & \textbf{70.50} & \textbf{88.25} & 52.75 \\
\rowcolor{gray!15} $\Delta$ (vs Qwen2.5-VL 7B) & {\scriptsize\textcolor{green!60!black}{$\uparrow 7.33$}} & {\scriptsize\textcolor{green!60!black}{$\uparrow 5.22$}} & {\scriptsize\textcolor{green!60!black}{$\uparrow 10.53$}} & {\scriptsize\textcolor{green!60!black}{$\uparrow 7.00$}} & {\scriptsize\textcolor{green!60!black}{$\uparrow 13.50$}} & {\scriptsize\textcolor{green!60!black}{$\uparrow 0.50$}} & {\scriptsize\textcolor{green!60!black}{$\uparrow 6.75$}} & {\scriptsize\textcolor{green!60!black}{$\uparrow 14.50$}} & {\scriptsize\textcolor{red!70!black}{$\downarrow 1.00$}} \\
\bottomrule
\end{tabular}
}
\end{table*}

\subsection{Experimental Settings}
\paragraph{\textup{\textbf{Training Data.}}}
In the SFT stage, following prior work, we use the Zebra-CoT~\cite{zebracot} dataset for cold-start training. Zebra-CoT covers four major task categories: scientific problems, 2D visual reasoning, 3D visual reasoning, and visual logic and strategy games, thereby providing diverse and logically coherent multimodal reasoning trajectories across a broad range of domains. In the RL stage, we curate and combine samples from the DeepEyes~\cite{deepeyes}, Thyme~\cite{thyme} and CodeDance~\cite{codedance} datasets to construct the training set.

\paragraph{\textup{\textbf{Benchmarks.}}}
We report results on three high-resolution benchmarks targeting ultra-high-definition perception and visual search: HRBench (4K \& 8K)~\cite{HRBench} and V*~\cite{vstar}. Additionally, we evaluate on a robust suite of general and diagnostic VQA benchmarks to cover various specialized capabilities: MMStar~\cite{mmstar} (visual dependency), MMVP~\cite{MMVP} (fine-grained visual patterns), SeedBench2Plus~\cite{li2024seed} (multimodal reasoning), BLINK~\cite{fu2024blink} (core visual perception), and HallusionBench~\cite{guan2024hallusionbench} (illusion and hallucination diagnostics).

\paragraph{\textup{\textbf{Baselines.}}}
To rigorously evaluate our approach, we benchmark it against four categories of representative baselines: (1) \textbf{Proprietary Frontier MLLMs}, including GPT-4o~\cite{gpt4o} and Gemini-3-Flash~\cite{gemini3flash2025}; (2) \textbf{Open-source General MLLMs}, including LLaVA-OneVision-7B~\cite{llava-onevision} and Qwen2.5-VL-7B~\cite{Qwen2.5-VL}; (3) \textbf{Thinking-with-Images Agent Models}, including ZoomEye~\cite{zoomeye}, DeepEyes~\cite{deepeyes}, DeepEyesV2~\cite{deepeyesv2}, and Thyme~\cite{thyme}; (4) \textbf{Visual Latent Reasoning Models}, including LVR~\cite{lvr-latent}, Laser~\cite{Laser}, Monet~\cite{monet}, and SkiLa~\cite{sketch}.

\paragraph{\textup{\textbf{Implementation Details.}}}
In the SFT stage, we limit training to a single epoch to mitigate overfitting. The learning rate is set to $10^{-5}$, the per-GPU batch size to 1, and the gradient accumulation steps to 16. Following this cold-start phase, we continue with DePO-based reinforcement learning, reducing the learning rate to $10^{-6}$. More implementation details can be found in supplemental materials.

\begin{table}[htbp]
\centering
\caption{Results on Multimodal VQA, Reasoning and Hallucination Benchmarks. The best-performing latent reasoning model for each dataset
is highlighted in \textbf{bold}.}
\label{tab:VQA_hall}
\resizebox{\textwidth}{!}{%
\begin{tabular}{l|cccc|c}
\toprule
\textbf{Methods} & \textbf{MMVP} & \textbf{MMStar} & \textbf{SeedBench2Plus} & \textbf{BLINK} & \textbf{HallusionBench} \\
\midrule
\rowcolor{blue!12} \multicolumn{6}{c}{\textit{Open-Source Model}} \\
Qwen2.5-VL-7B~\cite{Qwen2.5-VL} & 65.67 & 59.70 & 65.31 & 53.60 & 56.57 \\
LLaVA-OneVision~\cite{llava-onevision} & 73.00 & 59.13 & 61.22 & 49.34 & 51.10 \\
InternVL3.5-8B~\cite{internvl3.5} & 57.67 & 53.33 & 69.78 & 54.81 & 56.15 \\
\midrule
\rowcolor{violet!12} \multicolumn{6}{c}{\textit{Thinking-with-Images Agent Model}} \\
ZoomEye~\cite{zoomeye} & 72.67 & 63.20 & 70.27 & 55.55 & 71.08 \\
Thyme~\cite{thyme} & - & 65.90 & - & 56.10 & 55.60 \\
DeepEyes~\cite{deepeyes} & 70.00 & 58.73 & 69.08 & 51.08 & 62.57 \\
\midrule
\rowcolor{purple!10} \multicolumn{6}{c}{\textit{Visual-Latent Model}} \\
LVR~\cite{lvr-latent} & 64.00 & 57.93 & 47.39 & 53.60 & 65.19 \\
Laser~\cite{Laser} & 72.00 & 60.27 & 70.05 & 56.92 & \textbf{67.72} \\
Monet~\cite{monet} & 68.00 & 60.33 & 65.88 & 50.71 & 56.36 \\
\textbf{HyLaR-SFT (Ours)} & 71.00 & 60.47 & 69.39 & 54.50 & 60.23 \\
\textbf{HyLaR-7B (Ours)} & \textbf{73.67} & \textbf{62.00} & \textbf{70.32} & \textbf{57.14} & 63.68 \\
\rowcolor{purple!6} $\Delta$ (vs Qwen2.5-VL 7B) & {\scriptsize\textcolor{blue!70!black}{$\uparrow 8.00$}} & {\scriptsize\textcolor{blue!70!black}{$\uparrow 2.30$}} & {\scriptsize\textcolor{blue!70!black}{$\uparrow 5.01$}} & {\scriptsize\textcolor{blue!70!black}{$\uparrow 3.54$}} & {\scriptsize\textcolor{blue!70!black}{$\uparrow 7.11$}} \\
\bottomrule
\end{tabular}
}
\end{table}

\subsection{Main Results}

\paragraph{\textup{\textbf{High-Resolution Image Perception and Visual Search Benchmarks.}}}
In the benchmarks reported in Table~\ref{tab:main_benchmark}, the target regions occupy an extremely small fraction of ultra high resolution images, typically only 100 to 200 pixels. Accurately localizing such tiny targets in large-scale, visually cluttered imagery poses not only a substantial perceptual challenge for large models but also readily triggers severe visual hallucinations. As shown in the table, our model achieves significant performance gains across all three high-resolution benchmarks, and on some tasks it matches or surpasses Agent-based models that rely on external tools. Concretely, compared with the baseline Qwen2.5-VL-7B~\cite{Qwen2.5-VL}, our model improves by 7.33\% and 7.00\% on V*~\cite{vstar} and HRBench-4K~\cite{HRBench}, respectively. Notably, our model achieves state-of-the-art performance in visual latent space reasoning. On V*~\cite{vstar}, it outperforms SkiLa (78.53\%)~\cite{sketch} and Monet (80.10\%)~\cite{monet} by margins of 5.24\% and 3.67\%, respectively. This superiority extends to other high-resolution evaluations~\cite{HRBench}: on HRBench-4K, our model exceeds SkiLa (72.12\%) and Monet (67.37\%) by 2.88\% and 7.63\%; similarly, on HRBench-8K, it surpasses their respective scores of 66.50\% and 64.37\% by 4.00\% and 6.13\%.

\paragraph{\textup{\textbf{General VQA and Hallucination Benchmarks.}}}
Beyond fundamental capabilities, we further validate the effectiveness of our model on general VQA and hallucination evaluation benchmarks. Experimental results in Table~\ref{tab:VQA_hall} demonstrate that HyLaR substantially improves VQA accuracy while significantly mitigating hallucination phenomena in model generation. This can be attributed to HyLaR's capability of leveraging latent space representation guidance to enable deep exploration and fine-grained decomposition of specific local regions during inference, thereby achieving precise verification of visual content. These findings indicate that HyLaR successfully integrates high-resolution perception with an efficient verification mechanism, providing robust guarantees for the reliability of multimodal models.

\begin{wraptable}{r}{0.42\textwidth}
  \caption{Inference efficiency (normalized to Qwen2.5-VL-7B).}
  \label{tab:efficiency}
  \centering
  \scriptsize
  \setlength{\tabcolsep}{4pt}
  \begin{tabular}{@{}lcc@{}}
    \toprule
    Method & Latency & Memory \\
    \midrule
    Qwen2.5-VL-7B        & $1.00\times$ & $1.00\times$ \\
    DeepEyes (multi-turn) & $2.28\times$ & $3.64\times$ \\
    HyLaR ($N{=}16$)     & $\mathbf{0.96\times}$ & $0.98\times$ \\
    \bottomrule
  \end{tabular}
\end{wraptable}

\paragraph{\textup{\textbf{Efficiency Analysis.}}}
Benchmarked on an H20 GPU, HyLaR adds just 16 latent tokens and discards the SFT canvas extractor at inference, incurring zero architectural overhead. Unlike tool-based methods requiring detokenization, re-encoding, and external API calls, it reasons within a single forward pass. As shown in Table~\ref{tab:efficiency}, HyLaR runs at $0.96\times$ latency and $0.98\times$ memory of the Qwen2.5-VL-7B long-CoT baseline, whereas DeepEyes costs $2.28\times$ and $3.64\times$.

\paragraph{\textup{\textbf{Scalability to Smaller Model Sizes.}}}
To validate generalizability beyond the 7B scale, we evaluate a 3B variant 
against its backbone and the only available latent reasoning baseline at this scale. As shown in Table~\ref{tab:3b_results}, HyLaR-3B consistently outperforms both Qwen2.5-VL-3B and LVR-3B across all benchmarks, confirming that our framework scales effectively without loss of advantage.

\begin{table}[htbp]
\caption{Results of 3B-parameter models. \textbf{Bold} denotes the best latent reasoning model at the 3B scale.}
\label{tab:3b_results}
\centering
\scriptsize
\setlength{\tabcolsep}{4pt}
\begin{tabular}{@{}lcccccc@{}}
\toprule
\rowcolor{white}
Method & V$^*$ & HRBench-4K & HRBench-8K & MMVP & MMStar & SeedBench2Plus \\
\midrule
Qwen2.5-VL-3B   & \textbf{70.68} & 61.75 & 60.25 & 62.00 & 52.40 & 58.82 \\
LVR-3B          & 67.00 & -   & -   & 58.00 & -   & -   \\
HyLaR-3B (Ours) 
                & 68.58
                & \textbf{62.88} 
                & \textbf{61.25}
                & \textbf{63.00}
                & \textbf{54.20} 
                & \textbf{59.21} \\
\bottomrule
\end{tabular}
\end{table}

\subsection{Ablation Studies}
In this section, we conduct comprehensive ablation studies to evaluate the necessity and effectiveness of each component within our framework.

\begin{figure}[htbp]
  \centering
  \begin{subfigure}[b]{0.48\textwidth}
    \centering
    \includegraphics[width=\textwidth]{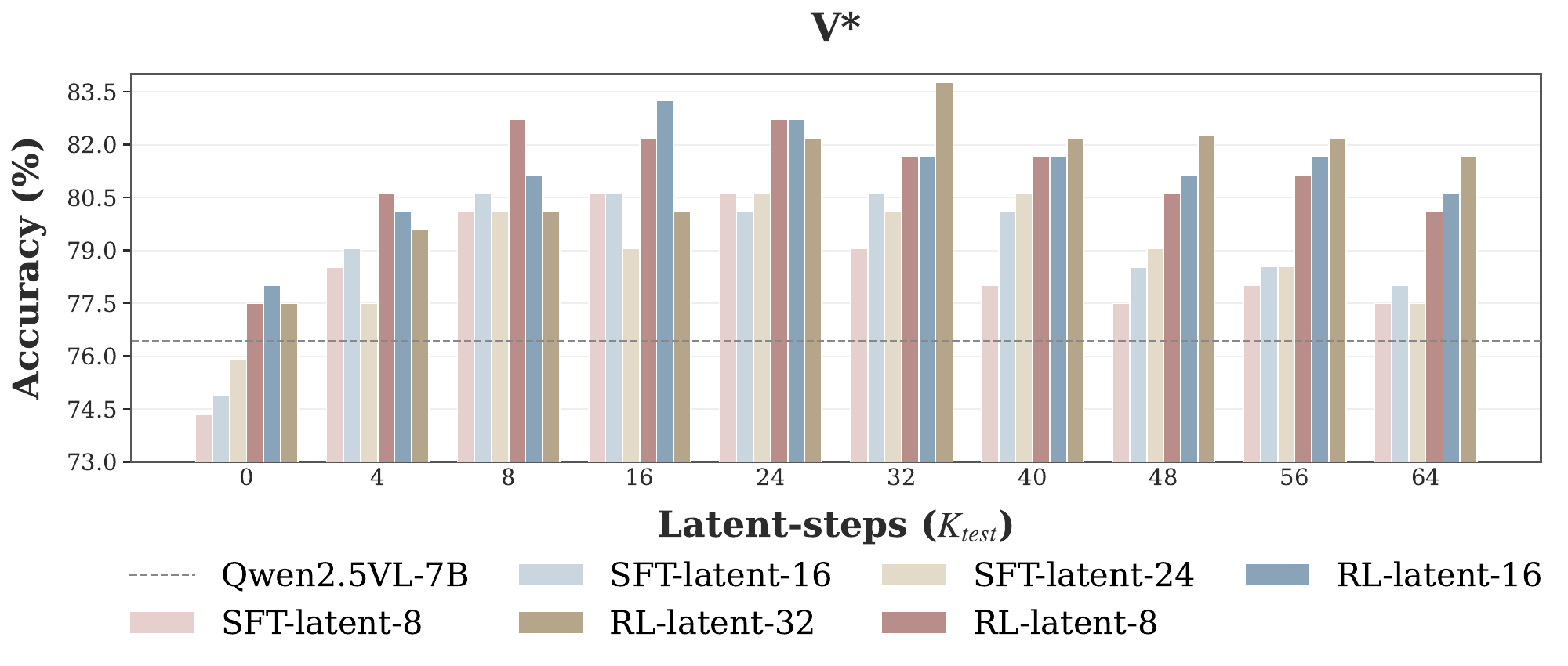}
    \label{fig:sub1}
  \end{subfigure}
  \hfill
  \begin{subfigure}[b]{0.48\textwidth}
    \centering
    \includegraphics[width=\textwidth]{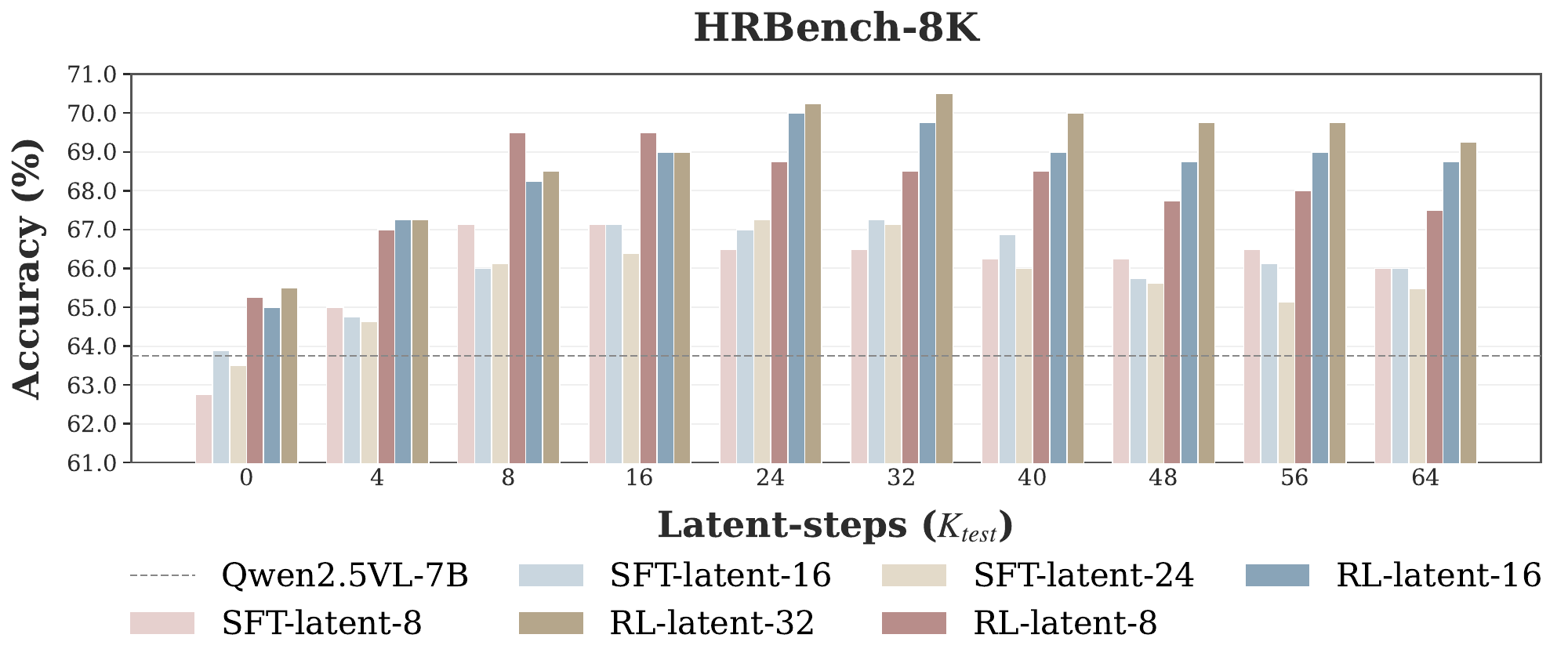}
    \label{fig:sub2}
  \end{subfigure}
  \hfill
  \caption{\textbf{Ablation on inference latent steps ($K_{\mathrm{test}}$).} SFT and RL models trained with varying horizons ($K_{\mathrm{train}}$), evaluated on V* and HRBench-8K. The dashed line marks the Qwen2.5-VL-7B baseline.}
  \label{fig:allthree}
\end{figure}

\paragraph{\textup{\textbf{Ablation on the Number of Latent Steps.}}}
To comprehensively understand the scaling behavior and generalization of latent reasoning, we investigate the interplay between training ($K_{\mathrm{train}}$) and inference ($K_{\mathrm{test}}$) reasoning steps across both SFT and RL stages. 
For the SFT stage, we train models with $K_{\mathrm{train}} \in \{8, 16, 24\}$. To further explore if RL can extrapolate reasoning beyond SFT priors, we initialize RL models from the SFT-8-steps checkpoint and optimize them with varying horizons: $K_{\mathrm{train}} \in \{8, 16, 32\}$. We systematically evaluate all models by varying $K_{\mathrm{test}}$ from 0 to 64 across two benchmarks: V* and HRBench-8K (as illustrated in Fig.~\ref{fig:allthree}). This unified setup reveals several key findings:
\textbf{(1) Latent reasoning consistently yields significant gains.} Across all benchmarks, introducing latent tokens ($K_{\mathrm{test}} > 0$) strictly outperforms the zero-latent baseline ($K_{\mathrm{test}} = 0$). For instance, on V*, SFT-latent-16 improves from 74.87\% ($K_{\mathrm{test}}=0$) to 80.63\% ($K_{\mathrm{test}}=16$), a substantial gain of \textbf{+5.76\%}. This demonstrates that latent tokens provide crucial ``thinking time'' for deeper implicit computation.
\textbf{(2) SFT models suffer from ``over-thinking'' degradation.} In the SFT stage, performance consistently degrades when $K_{\mathrm{test}}$ significantly exceeds $K_{\mathrm{train}}$. This indicates that excessive latent steps introduce noise or drift in hidden representations. Among SFT configurations, moderate training depth ($K_{\mathrm{train}}=16$) provides the most balanced trade-off between peak accuracy and generalization stability compared to $K_{\mathrm{train}}=8$ or $24$.
\textbf{(3) RL unlocks reasoning extrapolation and mitigates over-thinking.} Unlike SFT models that overfit to their training length, RL optimization significantly enhances length generalization. Models trained with RL, particularly with longer horizons ($K_{\mathrm{train}} \in \{16, 32\}$), maintain robust performance even when $K_{\mathrm{test}} \gg K_{\mathrm{train}}$, effectively mitigating the representation drift seen in SFT. Furthermore, RL pushes the peak accuracy beyond the upper bound of SFT, demonstrating that reinforcement learning can intrinsically align latent representations to better utilize extended inference budgets, an extrapolation capability that pure supervised learning fails to achieve.

\paragraph{\textup{\textbf{Ablation on Different RL Algorithms.}}}
To validate the effectiveness of our latent reasoning training approach, we conduct an ablation study by masking the intermediate thinking images in the SFT training data and performing standard SFT. As shown in Table~\ref{tab:ablation_rl}, the model trained with conventional SFT consistently underperforms our proposed method across various benchmarks, demonstrating the superiority of the latent approach. 
Furthermore, we compare DePO against three alternative optimization methods: GRPO~\cite{grpo}, DAPO~\cite{yu2025dapo}, and VLPO~\cite{monet}, with evaluations conducted on V*, HRBench (4K \& 8K), and MMVP.
The results demonstrate that DePO, with its hybrid optimization approach, achieves superior performance compared to these baseline RL algorithms. 
Notably, the models trained using GRPO and DAPO even showed a significant decline on the V* benchmark (from 80.63\% to 78.53\% and 79.06\%), while VLPO only achieves marginal improvements over HyLaR-SFT.
These compelling results highlight the remarkable effectiveness of DePO in dramatically advancing the performance of visual latent reasoning models on fine-grained feature understanding tasks.

\begin{table}[htbp]
\centering
\small
\caption{\textbf{Ablation on different RL algorithms.} All RL methods start from the same HyLaR-SFT checkpoint trained with 8 latent steps.}
\label{tab:ablation_rl}
\begin{tabular}{l cccc c}
\toprule
\textbf{Methods} & \textbf{V*} & \textbf{HRBench-4K} & \textbf{HRBench-8K} & \textbf{MMVP} & \textbf{Avg.} \\
\midrule
\multicolumn{6}{l}{\textit{Supervised Fine-tuning}} \\
\quad Text-only SFT & 69.11 & 65.88 & 61.00 & 70.20 & 66.55 \\
\rowcolor{purple!10} 
\quad HyLaR-SFT & 80.63 & 71.50 & 67.19 & 71.00 & 72.58 \\
\midrule
\multicolumn{6}{l}{\textit{Reinforcement Learning (from HyLaR-SFT)}} \\
\quad + GRPO~\cite{grpo} & 78.53 & 71.50 & 67.50 & 71.40 & 72.23 \\
\quad + DAPO~\cite{yu2025dapo} & 79.06 & 71.75 & 68.00 & 70.20 & 72.25 \\
\quad + VLPO~\cite{monet} & 80.10 & 72.84 & 68.00 & 69.67 & 72.65 \\
\rowcolor{blue!8}
\quad + DePO (Ours) & \textbf{83.77} & \textbf{75.00} & \textbf{70.50} & \textbf{73.67} & \textbf{75.74} \\
\bottomrule
\end{tabular}
\end{table}

\paragraph{\textup{\textbf{Ablation on DePO Hyper-parameters.}}}
We ablate two key hyper-parameters in DePO's decoupled policy loss
$\mathcal{L}_\text{PPO} = \mathcal{L}_\text{tok} + \alpha \cdot \mathcal{L}_\text{lat}$:
the latent loss weight~$\alpha$ and the latent clipping ratio~$[\epsilon_l^\text{lat}, \epsilon_h^\text{lat}]$,
evaluating on V*, HRBench-8K, and MMStar.

\noindent\textbf{Latent loss weight $\alpha$.}\;
As shown in Table~\ref{tab:ablation_alpha}, $\alpha=0$ (no RL signal to latents) consistently underperforms across all three benchmarks, confirming that latent tokens must be actively optimized. Increasing $\alpha$ to $2.0$ also degrades results; although our vMF formulation naturally mitigates sampling variance, an excessively large weight causes the continuous latent updates to disproportionately overwhelm the discrete text gradients, disrupting the delicate balance of the hybrid policy. Setting $\alpha=0.5$ strikes the best balance between sufficient latent optimization and training stability.

\noindent\textbf{Latent clipping ratio $[\epsilon_l^\text{lat}, \epsilon_h^\text{lat}]$.}\;
Table~\ref{tab:ablation_epsilon} shows that applying a clipping ratio equivalent to the lower bound of the text tokens ($\epsilon_l^{\mathrm{lat}} = \epsilon_h^{\mathrm{lat}} = \epsilon_l^{\mathrm{tok}} = 0.2$) leads to clear performance drops, since the continuous density ratio inherently exhibits different variance characteristics and requires a tighter trust region. Conversely, overly aggressive clipping ($\epsilon_l^\text{lat}= \epsilon_h^\text{lat}=0.01$) suppresses useful policy updates. Setting $\epsilon_l^\text{lat}= \epsilon_h^\text{lat}=0.05$ achieves the best overall performance across V*, HRBench-8K, and MMStar, validating the necessity of decoupled clipping.

\paragraph{\textup{\textbf{Ablation on Latent Distribution Assumptions.}}}

Our RL framework models continuous latent embeddings with the \textbf{von Mises-Fisher (vMF) distribution}, optimizing the policy via cosine similarity (directional alignment) between generated and rollout embeddings. To justify this choice, we replace vMF with a standard \textbf{Gaussian distribution}, under which embeddings are sampled around the policy output $\mathbf{h}_{i,t}^\theta$ and the probability ratio $r_{i,t}(\theta)$ is driven by Euclidean ($\ell_2$) distance rather than angular similarity:
\begin{equation}r_{i,t}^{\text{Gaussian}}(\theta)  = \exp \left( - \frac{1}{2\sigma^2} \|\mathbf{h}_{i,t}^{\text{old}} - \mathbf{h}_{i,t}^\theta\|^2 \right)
\label{eq:gaussian_ablation}
\end{equation}
where $\sigma$ controls the variance, tuned so the initial gradient scales match our vMF setting for fair comparison. Results are summarized in Table~\ref{tab:distribution_ablation}.

\begin{table}[t]
\begin{minipage}[t]{0.48\linewidth}
\centering
\caption{
  \textbf{Ablation on latent loss weight $\alpha$.}
  We vary the latent loss weight $\alpha$ in
  $\mathcal{L}_\text{PPO} = \mathcal{L}_\text{tok} + \alpha \cdot \mathcal{L}_\text{lat}$
  while keeping all other hyper-parameters fixed.
  During evaluation, we set the latent step to 32.
}
\label{tab:ablation_alpha}
\begin{tabular}{c|cccc}
\toprule
$\alpha$ & \textbf{V*} & \textbf{HRBench-8K} & \textbf{MMStar} \\
\midrule
0.0  & 80.63  & 68.63  & 60.17   \\
0.2  & 81.15  & 69.38  & 60.04   \\
0.3  & 81.68  & 70.00  & 60.71   \\
\rowcolor{blue!8}
\textbf{0.5}  & \textbf{83.77}  & \textbf{70.50}  & \textbf{62.00} \\
1.0  & 83.25  & 70.50  & 61.51   \\
2.0  & 80.63  & 69.50  & 61.04   \\
\bottomrule
\end{tabular}
\end{minipage}
\hfill
\begin{minipage}[t]{0.48\linewidth}
\centering
\caption{
  \textbf{Effect of latent clipping ratio $\epsilon^{\mathrm{lat}}$.}
  We set $\epsilon_l^\text{lat} = \epsilon_h^\text{lat} = \epsilon^\text{lat}$
  (symmetric clipping).
  Token clipping is fixed at $(\epsilon_l^\text{tok}, \epsilon_h^\text{tok}) = (0.20, 0.28)$.
  During evaluation, we set the latent step to 32.
}
\label{tab:ablation_epsilon}
\begin{tabular}{c|cccc}
\toprule
$\epsilon^{\mathrm{lat}}$ & \textbf{V*} & \textbf{HRBench-8K} & \textbf{MMStar} \\
\midrule
0.0  & 79.58  & 69.37  & 61.00   \\
0.01  & 82.72  & 70.00  & 61.40   \\
0.025  & 82.72  & \textbf{70.75}  & 61.67   \\
\rowcolor{blue!8}
\textbf{0.05}  & \textbf{83.77}  & 70.50  & \textbf{62.00} \\
0.1  & 83.77  & 70.00  & 61.47   \\
0.2  & 80.10  & 69.00  & 60.87   \\
\bottomrule
\end{tabular}
\end{minipage}
\end{table}

\begin{table}[htbp]  
  \centering  
  \caption{Ablation on the distribution assumptions for latent embeddings during RL. The vMF distribution (our main method) consistently outperforms the Gaussian variant across all benchmarks.}  
  \label{tab:distribution_ablation}  
  \resizebox{\textwidth}{!}{  
    \begin{tabular}{lcccccc}  
    \toprule  
    \textbf{Methods / Distribution Assumption} & \textbf{V*} & \textbf{HRBench-4K} & \textbf{HRBench-8K} & \textbf{SeedBench2Plus} & \textbf{MMStar} & \textbf{MMVP} \\  
    \midrule  
    Baseline (SFT only, w/o RL) & 80.63 & 71.50 & 67.19 & 65.31 & 60.47 & 71.00 \\  
    \rowcolor{purple!10} 
    RL with \textbf{Gaussian} ($\sigma=10.0$) & 83.25 & 74.00 & 69.50 & 67.28 & 61.27 & 72.33 \\  
    \rowcolor{blue!10}   
    RL with \textbf{vMF} (Ours, $\kappa=0.01$) & \textbf{83.77} & \textbf{75.00} & \textbf{70.50} & \textbf{70.32} & \textbf{62.00} & \textbf{73.67} \\  
    \bottomrule  
    \end{tabular}%
  }  
\end{table}

Based on the results in Table~\ref{tab:distribution_ablation}, we draw the following conclusions:
\textbf{(1) Directional alignment is more effective than spatial distance for latent space.} While the Gaussian assumption improves upon the SFT baseline, it consistently underperforms our vMF-based approach across all benchmarks. In high-dimensional spaces (such as the hidden states of LLMs), semantic information is primarily encoded in the direction of the vectors rather than their magnitude. By explicitly optimizing cosine similarity, the vMF distribution aligns perfectly with the nature of dense representations, leading to more meaningful policy updates.
\textbf{(2) vMF provides natural regularization against norm explosion.} Optimizing the Euclidean distance under the Gaussian assumption can lead to training instability. Specifically, to maximize the advantage $\hat{A}_{i,t}$, the policy might trivially increase the magnitude (norm) of the latent embeddings to push them further apart, rather than learning better semantic representations. The vMF distribution, which inherently operates on the unit hypersphere, acts as a natural regularizer. It forces the model to focus solely on angular updates, thereby stabilizing the RL training process.

\section{Conclusion}
\label{sec:conclusion}
In this work, we propose HyLaR (Hybrid Latent Reasoning), a novel framework designed to overcome early semantic collapse in MLLMs by seamlessly interleaving discrete textual logic with continuous visual working memory. To effectively optimize this hybrid discrete-continuous action space, we introduce Decoupled Policy Optimization (DePO). By identifying the geometric and variance mismatches inherent in standard RL, DePO leverages von Mises-Fisher (vMF) spherical modeling and decoupled trust-region clipping to achieve exact, stable, and geometrically rigorous latent policy optimization. Extensive experiments demonstrate that HyLaR significantly outperforms existing text-only and tool-augmented paradigms on fine-grained perception and reasoning benchmarks, exhibiting remarkable robustness to visual hallucinations. In future work, we aim to extend this hybrid RL paradigm to open-ended agentic environments and explore scaling laws for longer-horizon latent planning.


\bibliographystyle{splncs04}
\bibliography{main}

\title{HyLaR: Hybrid Latent Reasoning with Decoupled Policy Optimization \\ 
\Large \textit{(Supplementary Material)}} 

\titlerunning{Hybrid Latent Reasoning with Decoupled Policy Optimization}

\author{Tao Cheng\inst{1} \and Shi-Zhe Chen\inst{1} \and Hao Zhang\inst{1} \and Yixin Qin\inst{1} \\ Jinwen Luo\inst{2} \and Zheng Wei\inst{1}}

\authorrunning{T.~Cheng et al.}

\institute{$^1$ Tencent PCG \quad $^2$ Tencent CSIG \\ \email{\{elvistcheng,shizhechen,jeffhzhang,wadeqin,jamsluo,hemingwei\}@tencent.com}}

\maketitle
\setcounter{equation}{0}
\setcounter{figure}{0}
\setcounter{table}{0}
\renewcommand{\theequation}{S\arabic{equation}}
\renewcommand{\thefigure}{S\arabic{figure}}
\renewcommand{\thetable}{S\arabic{table}}

\section{Derivation of the von Mises-Fisher (vMF) Policy Optimization}
\label{sec:vmf_derivation}

Gaussian policies are widely used in continuous control, but they implicitly assume a flat Euclidean action space. In contrast, latent representations in modern Multimodal Large Language Models (MLLMs) are typically processed by normalization layers (e.g., RMSNorm), making their geometry primarily directional rather than purely Euclidean. This mismatch suggests that policy optimization in latent space should respect hyperspherical structure and angular similarity.

To capture this geometry, we model the latent policy using the von Mises--Fisher (vMF) distribution, the canonical probability distribution on the unit hypersphere $\mathbb{S}^{D-1}$. Importantly, in our formulation, the vMF distribution is \emph{not} used to sample stochastic actions during rollout. Instead, it serves as a directional density model for evaluating how well the current policy aligns with the reference latent direction produced by the old policy. Concretely, the optimization target is the old policy mode itself.

\subsection{Mode-Referenced Density Ratio}

For a state $s_t$, let the current policy produce a normalized latent direction
$\boldsymbol{\mu}_t^\theta \in \mathbb{S}^{D-1}$,
and define the corresponding vMF density
\begin{equation}
p_\theta(a_t \mid s_t)=
C_D(\kappa)\exp\!\big(\kappa (\boldsymbol{\mu}_t^\theta)^\top \mathbf{\tilde z}_t\big),
\qquad a_t \in \mathbb{S}^{D-1},
\end{equation}
where $\kappa > 0$ is the concentration parameter and $C_D(\kappa)$ is the normalization constant.
During rollout, we store the old policy direction
$\mathbf{\tilde z}_t = \boldsymbol{\mu}_t^{\theta_{\mathrm{old}}}$,
which is the mode of the old vMF policy. In the subsequent optimization step, we evaluate both the old and new policies at this same reference point $a_t = \mathbf{\tilde z}_t$. Therefore, under the old policy,
\begin{equation}
\pi_{\theta_{\mathrm{old}}}(a_t \mid s_t)=
\pi_{\theta_{\mathrm{old}}}(\mathbf{\tilde z}_t \mid s_t)=
C_D(\kappa)\exp\!\big(\kappa (\mathbf{\tilde z}_t)^\top \mathbf{\tilde z}_t\big)=
C_D(\kappa)\exp(\kappa),
\end{equation}
Under the new policy, the density at the same reference point is
\begin{equation}
\pi_\theta(a_t \mid s_t)=
\pi_\theta(\mathbf{\tilde z}_t \mid s_t)=
C_D(\kappa)\exp\!\big(\kappa (\boldsymbol{\mu}_t^\theta)^\top \mathbf{\tilde z}_t\big),
\end{equation}
Hence, the mode-referenced density ratio is
\begin{equation}
r_t=
\frac{\pi_\theta(a_t \mid s_t)}{\pi_{\theta_{\mathrm{old}}}(a_t \mid s_t)}=
\exp\!\Big(\kappa\big((\boldsymbol{\mu}_t^\theta)^\top \mathbf{\tilde z}_t - 1\big)\Big),
\end{equation}
Since both $\boldsymbol{\mu}_t^\theta$ and $\mathbf{\tilde z}_t$ are $\ell_2$-normalized, their inner product is exactly the cosine similarity:
\begin{equation}
(\boldsymbol{\mu}_t^\theta)^\top \mathbf{\tilde z}_t = \cos(\boldsymbol{\mu}_t^\theta, \mathbf{\tilde z}_t),
\end{equation}
Therefore, the log-density ratio is governed purely by angular deviation from the old policy mode:
\begin{equation}
\log r_t=
\kappa\big(\cos(\boldsymbol{\mu}_t^\theta, \mathbf{\tilde z}_t)-1\big),
\end{equation}
This formulation should be interpreted as a deterministic, mode-referenced policy update in latent space, rather than a standard importance-sampling ratio over sampled stochastic actions.

\subsection{Closed-form vMF KL Divergence}

We next derive the trust-region regularizer induced by the vMF geometry. Consider two vMF distributions with a shared concentration parameter $\kappa$:

\begin{equation}
p(x)=\mathrm{vMF}(\boldsymbol{\mu}_{\mathrm{new}}, \kappa),
\qquad
q(x)=\mathrm{vMF}(\boldsymbol{\mu}_{\mathrm{old}}, \kappa),
\end{equation}
Their Kullback--Leibler divergence is

\begin{equation}
D_{\mathrm{KL}}(p\|q)=
\mathbb{E}_{x\sim p}
\left[
\log
\frac{C_D(\kappa)\exp\!\big(\kappa \boldsymbol{\mu}_{\mathrm{new}}^\top x\big)}
     {C_D(\kappa)\exp\!\big(\kappa \boldsymbol{\mu}_{\mathrm{old}}^\top x\big)}
\right],
\end{equation}
The normalization constants cancel, giving
\begin{equation}
D_{\mathrm{KL}}(p\|q)=
\kappa(\boldsymbol{\boldsymbol{\mu}}_{\mathrm{new}}-\boldsymbol{\mu}_{\mathrm{old}})^\top \mathbb{E}_{x\sim p}[x],
\label{eq:DL_KL}
\end{equation}
For a vMF distribution, the mean satisfies
\begin{equation}
\mathbb{E}_{x\sim p}[x]=
A_D(\kappa)\boldsymbol{\mu}_{\mathrm{new}},
\end{equation}
where $A_D(\kappa)$ is the standard mean resultant length. Substituting this into Eq.~\eqref{eq:DL_KL}, we obtain
\begin{equation}
D_{\mathrm{KL}}(p\|q)=
\kappa A_D(\kappa)
\big(
\boldsymbol{\mu}_{\mathrm{new}}^\top \boldsymbol{\mu}_{\mathrm{new}}-
\boldsymbol{\mu}_{\mathrm{old}}^\top \boldsymbol{\mu}_{\mathrm{new}}
\big),
\end{equation}
Since $\|\boldsymbol{\mu}_{\mathrm{new}}\|_2=1$, this simplifies to
\begin{equation}
D_{\mathrm{KL}}(p\|q)=
\kappa A_D(\kappa)
\big(1-\boldsymbol{\mu}_{\mathrm{old}}^\top \boldsymbol{\mu}_{\mathrm{new}}\big).
\end{equation}

In our policy optimization setting, we identify
\begin{equation}
\boldsymbol{\mu}_{\mathrm{new}} = \boldsymbol{\mu}_t^\theta,
\qquad
\boldsymbol{\mu}_{\mathrm{old}} = \mathbf{\tilde z},
\end{equation}
because the stored rollout latent $\tilde z_t$ is precisely the old policy mode. Therefore, the latent KL regularizer becomes
\begin{equation}
\mathcal{L}_{\mathrm{KL}}^{\mathrm{lat}}=
\mathbf{W}_\kappa
\big(1-\cos(\boldsymbol{\mu}_t^\theta,\mathbf{\tilde z}_t)\big),
\qquad
\mathbf{W}_\kappa = \kappa A_D(\kappa),
\end{equation}
This result shows that, under hyperspherical normalization and shared concentration, the exact vMF KL divergence reduces to a scaled cosine distance between the current policy direction and the old policy mode. Thus, both the ratio term and the trust-region penalty are naturally aligned with angular geometry.

\subsection{Practical Relaxation of $\ell_2$-Normalization}

The derivations above assume that latent representations lie exactly on the unit hypersphere. This provides a clean geometric interpretation and yields closed-form directional objectives. However, in practical MLLM optimization, enforcing strict unit-norm normalization can be suboptimal.

First, pre-trained MLLMs often preserve meaningful activation magnitudes through normalization layers such as RMSNorm. Although these representations are directionally structured, their norms are typically not exactly $1$, and are often on the order of $\sqrt{D}$. Forcing them onto the unit sphere may distort the scale statistics inherited from pre-training and harm optimization stability.

Second, the latent norm itself can carry useful optimization signals. Writing the unnormalized inner product as
\begin{equation}
\mathbf{h}_t^\top \mathbf{\tilde z}_t=
\|\mathbf{h}_t\|_2 \, \|\mathbf{\tilde z}_t\|_2 \cos(\mathbf{h}_t,\mathbf{\tilde z}_t),
\end{equation}
we see that the norm modulates the sharpness of directional matching. This suggests that magnitude can act as a state-dependent concentration-like factor, allowing the policy to adaptively control update intensity across different reasoning steps.

Motivated by these observations, we retain the vMF derivation as the normalized theoretical foundation, but adopt an unnormalized surrogate in implementation. Specifically, we replace cosine-based terms by inner-product-based scores throughout the RL objective. In this relaxed formulation, the exact probabilistic interpretation as a vMF density ratio or KL divergence no longer strictly holds; instead, the resulting objective should be viewed as a geometry-preserving surrogate that inherits the directional preference of the vMF model while preserving magnitude information from the pre-trained latent space.

As detailed in Table~\ref{tab:norm_ablation}, several key observations emerge from this ablation:

\begin{itemize}
    \item \textbf{Consistent Improvement:} The unnormalized approach yields an average performance gain of $+2.87\%$ across all evaluated benchmarks, demonstrating the generalizability and robustness of our relaxed formulation.
    
    \item \textbf{Fine-Grained Tasks Benefit Most:} Notably, the performance gap is most pronounced on high-resolution, reasoning-intensive benchmarks such as HRBench-4K ($+3.50\%$) and HRBench-8K ($+4.00\%$). This highlights that allowing the policy to dynamically scale its latent magnitude is particularly crucial for tasks requiring deep, fine-grained visual feature analysis, where strictly normalized embeddings might prematurely bottleneck information capacity.
\end{itemize}

\begin{table}[htbp]
\centering
\small 
\caption{\textbf{Effect of $\ell_2$-Normalization Relaxation.} Performance comparison between normalized and unnormalized latent logit computation across various multimodal benchmarks. The unnormalized approach consistently yields superior performance, validating that relaxing the strict unit-norm constraint benefits the hybrid policy optimization. All metrics are reported in accuracy (\%).}
\label{tab:norm_ablation}
\begin{tabular}{l|ccccc|c}
\toprule
\textbf{Methods} & \textbf{V*} & \textbf{HRBench-4K} & \textbf{HRBench-8K} & \textbf{MMStar} & \textbf{MMVP} & \textbf{Avg.} \\
\midrule
Normalized      & 81.15 & 71.50 & 66.50 & 61.13 & 70.33 & 70.12 \\
Unnormalized    & \textbf{83.77} & \textbf{75.00} & \textbf{70.50} & \textbf{62.00} & \textbf{73.67} & \textbf{72.99} \\
\midrule
$\Delta$~(gains) & \textcolor{green!80!black}{+2.62} & \textcolor{green!80!black}{+3.50} & \textcolor{green!80!black}{+4.00} & \textcolor{green!80!black}{+0.87} & \textcolor{green!80!black}{+3.34} & \textcolor{green!80!black}{+2.87} \\
\bottomrule
\end{tabular}
\end{table}

Empirically, this relaxation consistently improves performance, indicating that practical policy optimization benefits from combining directional alignment with adaptive magnitude scaling.
    
\section{Implementation Details}

\subsection{Dataset}
We construct our SFT dataset from Zebra-CoT~\cite{zebracot}, a large-scale multimodal corpus with 182,384 samples that provides logically coherent interleaved text–image reasoning trajectories spanning four task categories: scientific problems, 2D visual reasoning, 3D visual reasoning, and visual logic and strategy games, each comprising multiple subdomains. We further filtered the dataset to remove samples unsuitable for constructing latent patterns, such as cases where overly complex intermediate reasoning processes make intermediate features difficult to reconstruct accurately. After this processing, our final training dataset comprises approximately 96K samples. The specific number of data points for each category is shown in Table~\ref{tab:sft_dataset}.

\begin{table}[htbp]
    \centering
    \caption{Detailed statistics of training data.}
    \label{tab:sft_dataset}
    \begin{tabular}{lcc}
        \toprule
        \textbf{Category} & \textbf{Count} & \textbf{Ratio (\%)} \\
        \midrule
        Visual Jigsaw & 21,899 & 22.77 \\
        Visual Search & 30,000 & 31.19 \\
        Geometry & 1,058 & 1.10 \\
        Checkers & 2,753 & 2.86 \\
        Chess & 20,483 & 21.29 \\
        Maze & 20,000 & 20.79 \\
        \midrule
        Total & 96,193 & 100.00 \\
        \bottomrule
    \end{tabular}
\end{table}

Our RL dataset is derived from three sources: DeepEyes~\cite{deepeyes}, Thyme~\cite{thyme}, and CodeDance~\cite{codedance}, with tasks primarily focused on the ``Think with Images'' paradigm and containing a large number of high-resolution input images. Due to substantial overlap among these three datasets, we first applied a similarity-based filtering method for deduplication, and then removed samples with open-ended (non-deterministic) answers to prevent misjudgment by the judge model during training. This process ultimately yielded a training dataset comprising 48,654 samples.

\subsection{Training}

\paragraph{\textup{\textbf{SFT Training.}}}
We implement our SFT based on the TRL library~\cite{trl}. The training is conducted on 8 NVIDIA H20 GPUs using the AdamW optimizer. To mitigate overfitting on the fine-tuning data, we limit the training to a single epoch. Detailed hyperparameters are summarized in Table~\ref{tab:hyperparameters_sft}.

\begin{table}[htbp]
  \centering
  \caption{Hyperparameters for SFT.}
  \label{tab:hyperparameters_sft}
  \begin{tabular}{lc}
    \toprule
    \textbf{Hyperparameter} & \textbf{Value} \\
    \midrule
    learning rate & $10^{-5}$ \\
    batch size & 1 \\
    gradient accumulation steps & 16 \\
    epoch & 1 \\
    latent size & 8 \\
    weight decay & 0.01 \\
    max pixels per image & $2000 \times 28 \times 28$ \\
    \bottomrule
  \end{tabular}
\end{table}

\paragraph{\textup{\textbf{RL Training.}}}
We implement our RL training using EasyR1~\cite{easyr1}, an open-source RL training framework for Multimodal LLMs. The training is conducted on 8 NVIDIA H20 GPUs, with hyperparameters summarized in Table~\ref{tab:hyperparameters_rl}. Notably, while the unnormalized latent magnitude intrinsically acts as a dynamic concentration parameter, we retain a small constant $\kappa = 0.01$ as a global base scale. This scaling factor constrains the initial logit variance, preventing premature entropy collapse during the early stages of policy exploration.

\textbf{Reward Design.} Our reward function comprises two components: \textit{accuracy reward} and \textit{format reward}. For accuracy evaluation, we employ a binary scoring mechanism: a reward of 1 is assigned if the model generates the correct answer, and 0 otherwise. GPT-5 serves as the evaluation model. For format reward, we introduce structural constraints to enforce a predefined reasoning paradigm. Specifically, the model is required to: (1) perform explicit reasoning within the \texttt{<think></think>} tags, (2) transition to implicit reasoning by generating the \texttt{<|canvas\_start|>} token, conclude the implicit reasoning phase with the \texttt{<|canvas\_end|>} token, and (3) encapsulate the final answer within the \texttt{<answer></answer>} tags. This explicit-implicit reasoning process can iterate alternately, enabling the model to synergistically leverage both textual chain-of-thought and visual latent representations. The model receives a format reward of 1 if its inference trace conforms to this explicit–implicit pattern and 0 otherwise. 

\textbf{Dynamic Rollout Filtering.} Our proposed DePO is built upon the Group Relative Policy Optimization (GRPO) algorithm, which computes the advantage of each sample by normalizing rewards within the same group:
\begin{equation}
  \hat{A}_{i} = \frac{r_i - \mathrm{mean}(\{r_j\}_{j=1}^{G})}{\mathrm{std}(\{r_j\}_{j=1}^{G}) + \epsilon}
\end{equation}
where $G$ denotes the rollout size and $r_i$ represents the reward of the $i$-th sample. A notable limitation of this group normalization mechanism is that when all samples within a group receive identical rewards (\emph{i.e.}, all correct or all incorrect), the standard deviation becomes zero, resulting in vanishing advantage values and zero policy gradients.

To address this issue and avoid wasting computational resources, we introduce a dynamic rollout filtering mechanism. Specifically, we discard any rollout group whose mean accuracy falls outside the range $[0.1, 0.9]$. This on-the-fly filtering ensures sufficient reward variance within each group, thereby maintaining effective gradient signals throughout the training process.

\begin{table}[htbp]
  \centering
  \caption{Hyperparameters for RL.}
  \label{tab:hyperparameters_rl}
  \begin{tabular}{lc}
    \toprule
    \textbf{Hyperparameter} & \textbf{Value} \\
    \midrule
    learning rate & $10^{-6}$ \\
    batch size & 64 \\
    weight decay & 0.01 \\
    rollout size ($G$) & 8 \\
    temperature & 0.9 \\
    max response length & 2048 \\
    $\kappa$ (global base scale) & 0.01 \\
    max pixels per image & $4000 \times 28 \times 28$ \\
    valid group accuracy & $[0.1, 0.9]$ \\
    \bottomrule
  \end{tabular}
\end{table}

\section{Case Studies}

\begin{figure}[htbp]
  \centering
  \includegraphics[width=\textwidth]{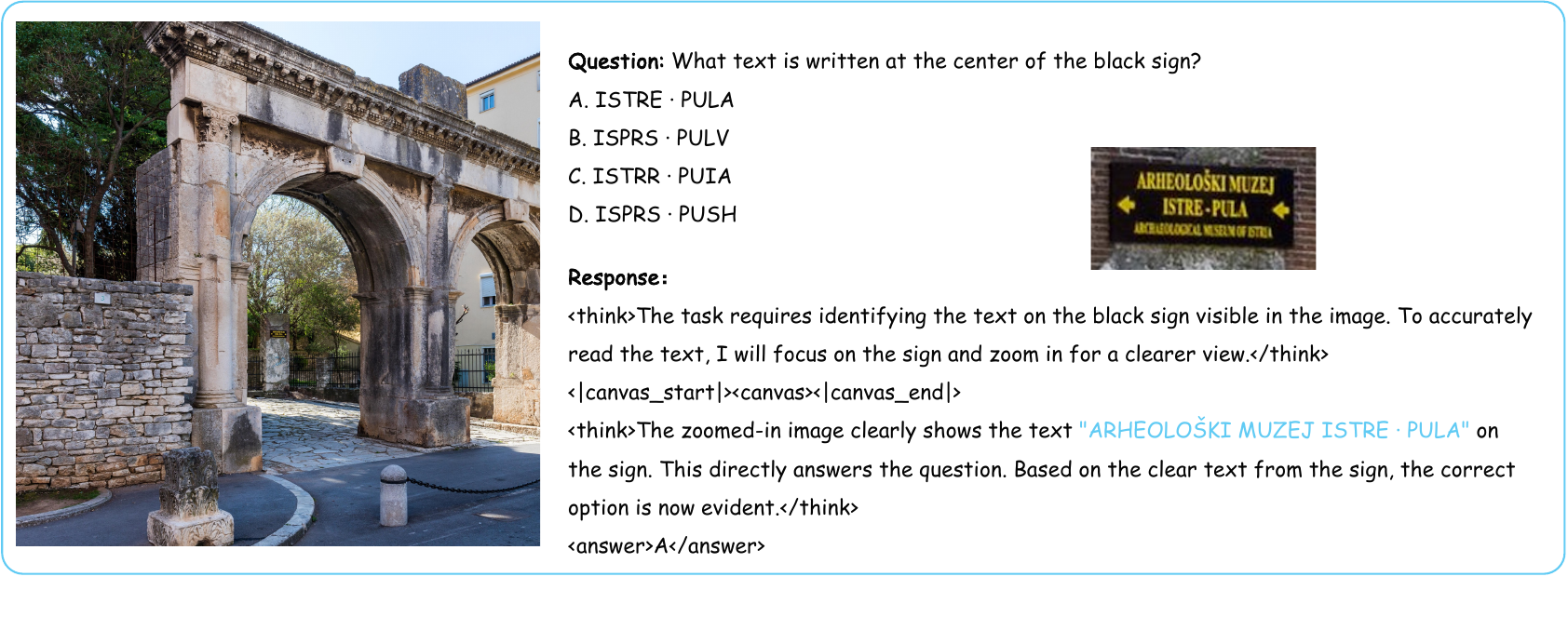}  
  \caption{Inference example: Fine-grain OCR.}
  \label{fig:ocr}                                
\end{figure}

\begin{figure}[htbp]
  \centering
  \includegraphics[width=\textwidth]{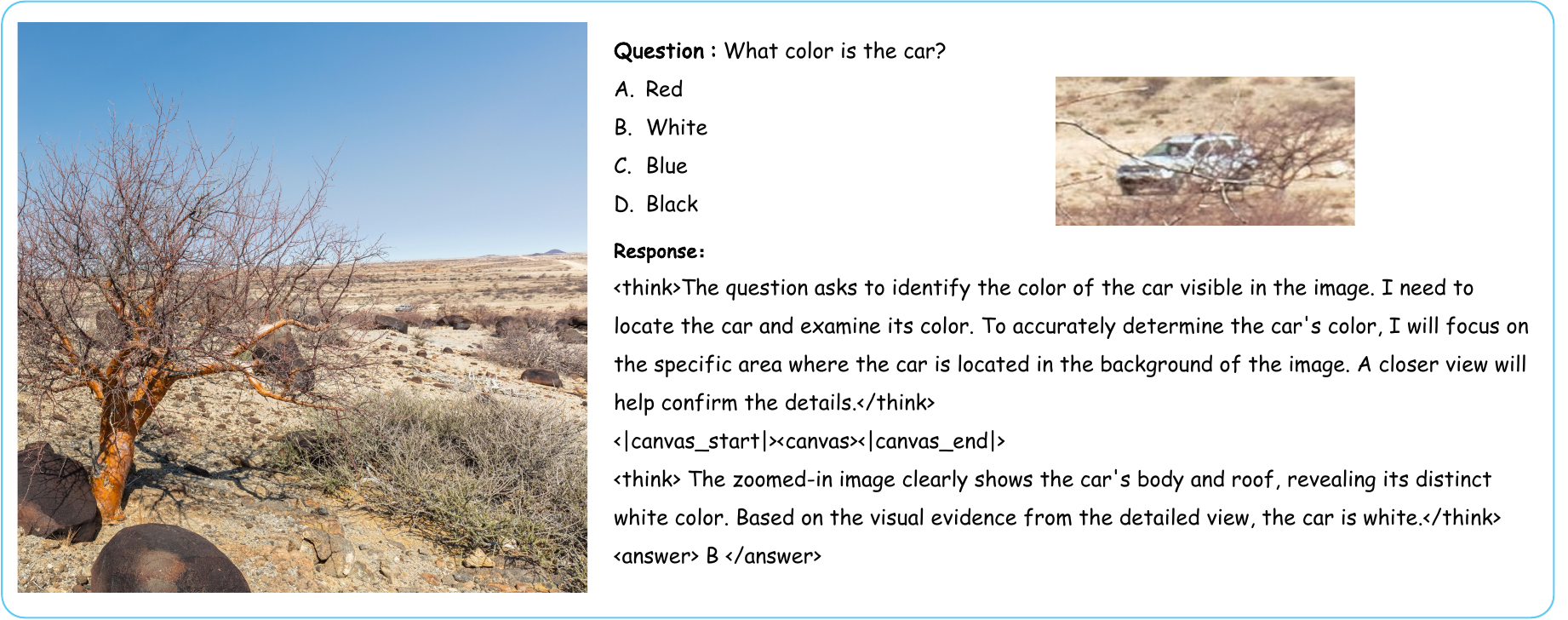}  
  \caption{Inference example: Target object search.}
  \label{fig:objsearch}                                
\end{figure}

\begin{figure}[htbp]
  \centering
  \includegraphics[width=\textwidth]{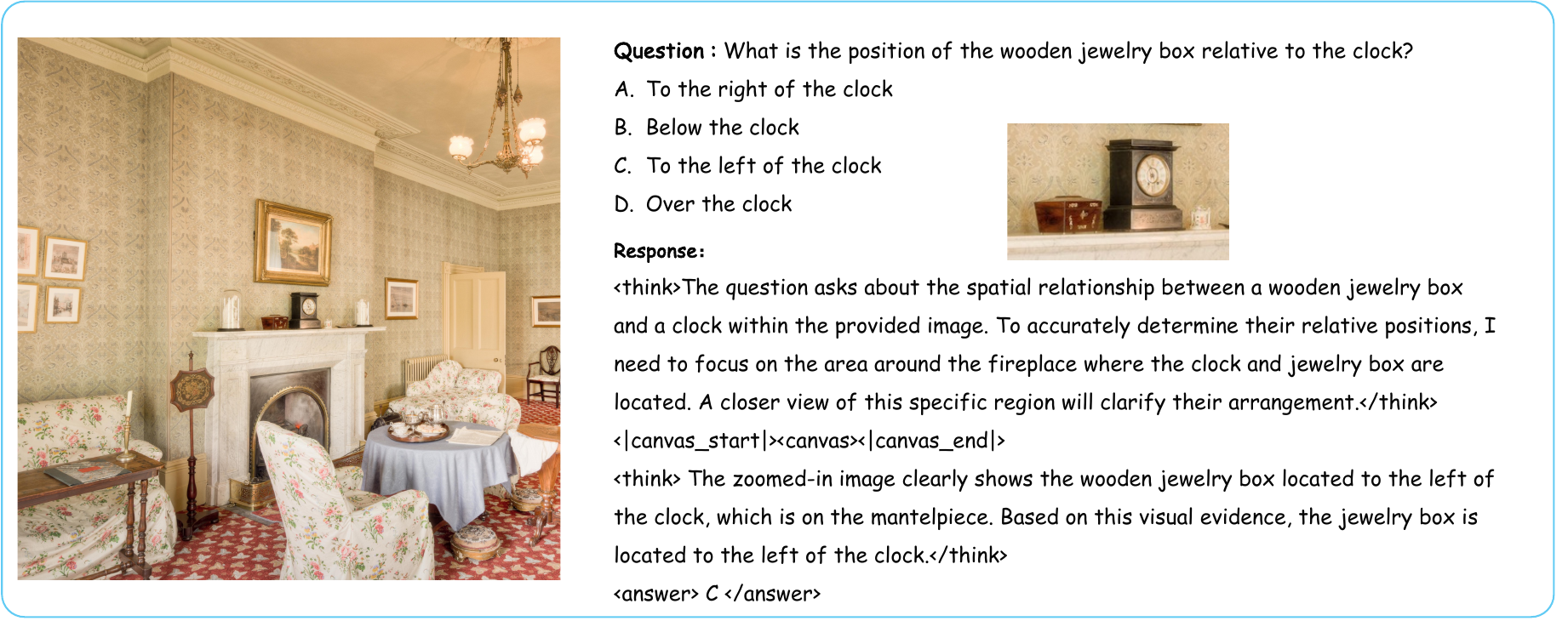}  
  \caption{Inference example: Spatial relationship.}
  \label{fig:spatialreasoning}                                
\end{figure}

\begin{figure}[htbp]
  \centering
  \includegraphics[width=\textwidth]{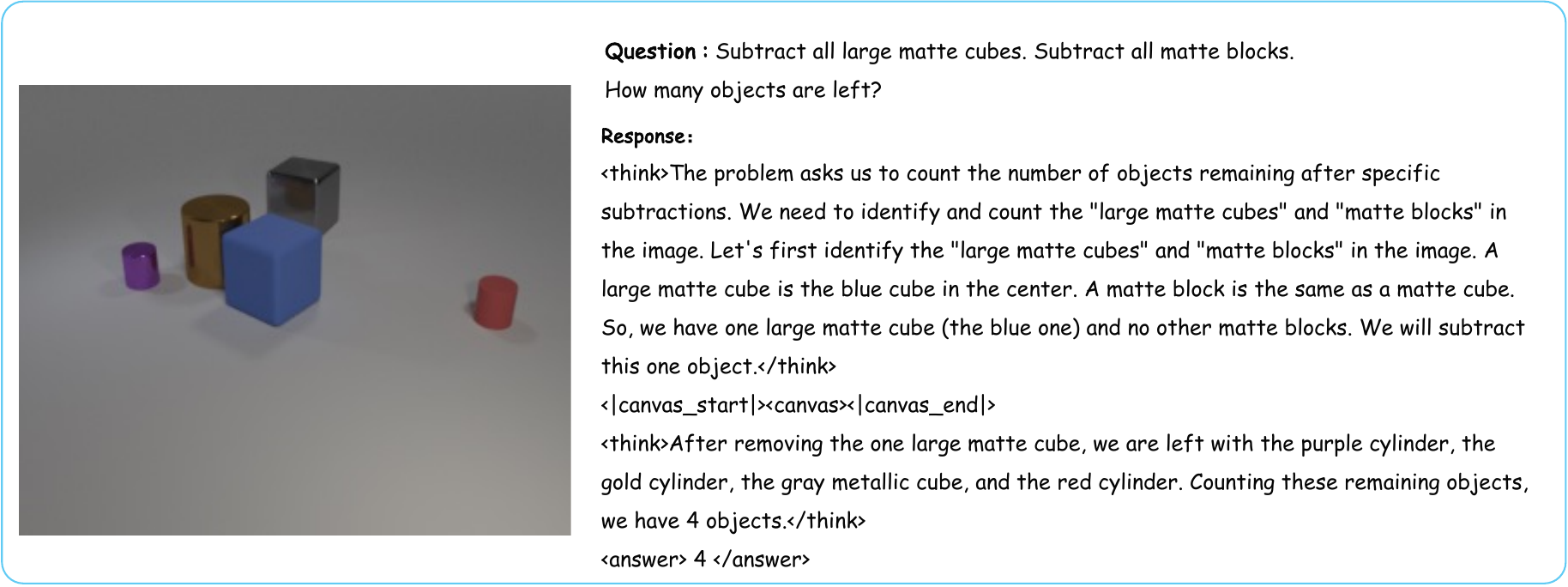}  
  \caption{Inference example: Counting three-dimensional objects.}
  \label{fig:counting3}                                
\end{figure}

\begin{figure}[htbp]
  \centering
  \includegraphics[width=\textwidth]{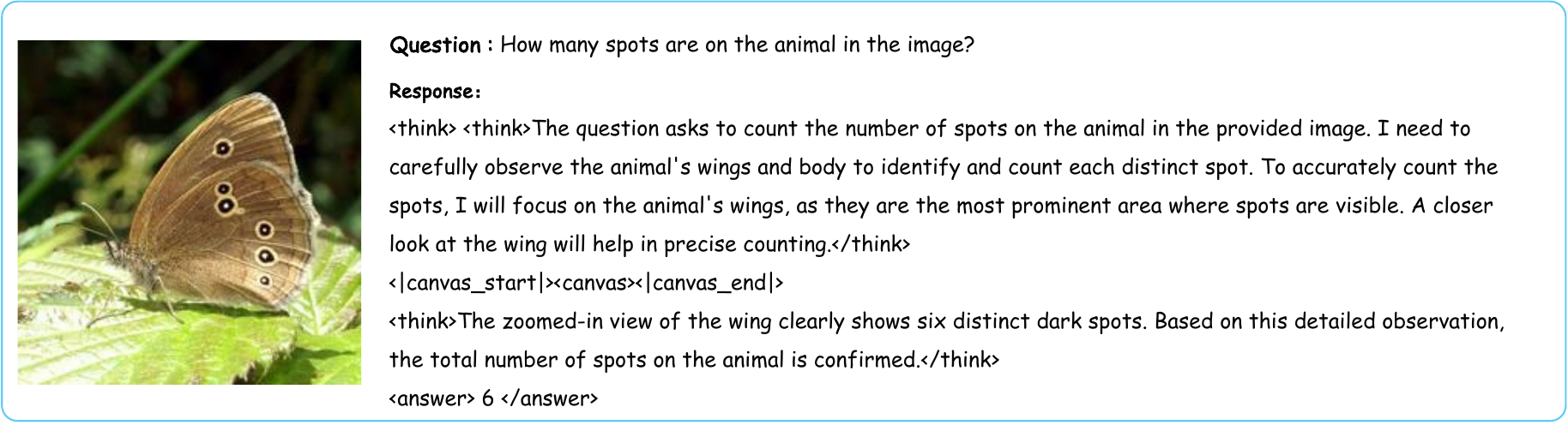}  
  \caption{Inference example: Counting two-dimensional objects.}
  \label{fig:counting2}                                
\end{figure}

\begin{figure}[htbp]
  \centering
  \includegraphics[width=\textwidth]{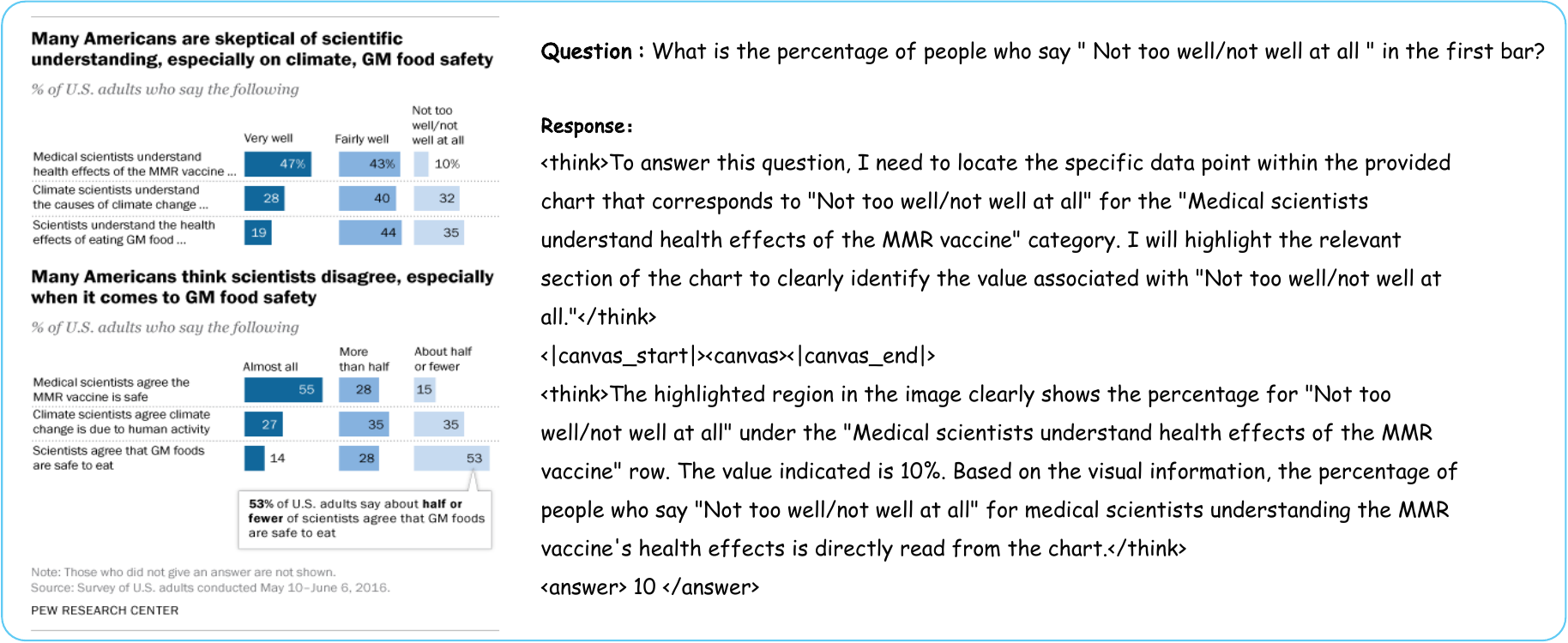}  
  \caption{Inference example: ChartQA.}
  \label{fig:chartqa}                                
\end{figure}

This section presents representative inference examples from HyLaR-7B to demonstrate its versatility across diverse tasks. We use \texttt{<|canvas\_start|>\allowbreak<canvas>\allowbreak<|canvas\_end|>} to denote the visual latent inference process. The selected examples encompass: fine-grained OCR (Fig.~\ref{fig:ocr}), target object search (Fig.~\ref{fig:objsearch}), real-world spatial reasoning (Fig.~\ref{fig:spatialreasoning}), three-dimensional counting (Fig.~\ref{fig:counting3}), two-dimensional counting (Fig.~\ref{fig:counting2}), and complex chart reasoning (Fig.~\ref{fig:chartqa}).
Specifically, across these diverse scenarios, we observe a consistent and interpretable reasoning pattern. When confronted with fine-grained visual queries, HyLaR proactively triggers the latent reasoning process (via the \texttt{<|canvas\_start|>} token) to allocate additional computational steps. This internal ``zooming'' or ``focusing'' mechanism allows the model to dynamically refine its visual working memory—such as locating the precise text on a distant sign (Fig.~\ref{fig:ocr}), isolating small target objects from complex backgrounds (Fig.~\ref{fig:objsearch}), or distinguishing clustered instances for accurate counting (Fig.~\ref{fig:counting3} and Fig.~\ref{fig:counting2})—before generating the final textual response. These cases empirically demonstrate that our HyLaR framework successfully aligns the continuous latent visual policy with complex discrete reasoning trajectories.

\section{Attention Visualizations}

\begin{figure}[htbp]
  \centering
  \begin{subfigure}[b]{\textwidth}
    \centering
    \includegraphics[width=\textwidth]{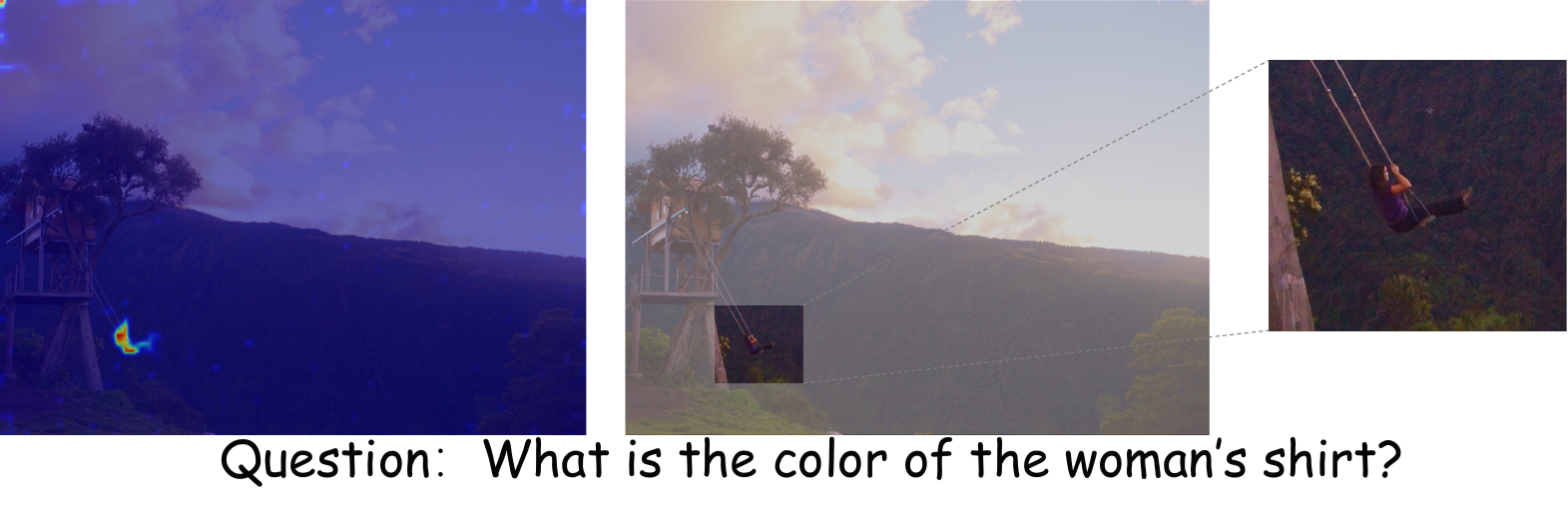}
    \caption{Object-attribute query (\emph{``What is the color of the woman's shirt?''}). The latent canvas tokens concentrate on the woman's torso region, bypassing background clutter for accurate color identification.}
    \label{fig:att_map_shirt}
  \end{subfigure}

  \begin{subfigure}[b]{\textwidth}
    \centering
    \includegraphics[width=\textwidth]{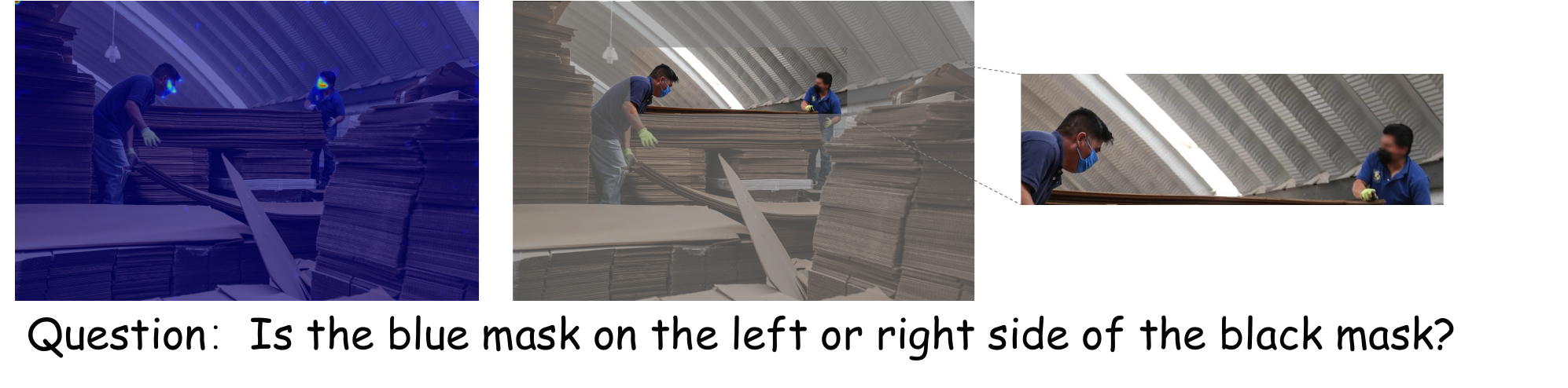}
    \caption{Spatial-relational query (\emph{``Is the blue mask on the left or right side of the black mask?''}). The latent canvas tokens jointly highlight both objects, capturing their relative spatial configuration for precise left-vs.-right judgment.}
    \label{fig:att_map_mask}
  \end{subfigure}

  \caption{Attention map visualizations of the latent canvas tokens on two query types.}
  \label{fig:att_map}
\end{figure}

The attention visualizations in Fig.~\ref{fig:att_map} confirm that the compressed latent canvas tokens function as salient semantic pointers. Rather than diffusely attending to the entire image, they sharply focus on fine-grained, reasoning-critical regions, such as the precise location of a distant object or a locally distinctive feature, thereby providing strict visual grounding for the subsequent textual reasoning steps.

\end{document}